\documentclass{article}

\usepackage{xcolor}
\definecolor{nicegreen}{rgb}{0.1,0.6,0.2}
\definecolor{niceblue}{rgb}{0.0,0.4,0.8}

\usepackage[final]{corl_2026} 

\usepackage{orcidlink}
\usepackage{hyperref}
\usepackage{xspace}
\usepackage{booktabs}
\usepackage{graphicx}
\usepackage{pifont}
\usepackage{array}
\usepackage{multirow}
\usepackage{amsmath}
\usepackage{tabularx}
\usepackage{float}
\usepackage{subcaption}
\usepackage{amsfonts}
\usepackage{amssymb}
\usepackage{makecell}
\usepackage{float}

\def\name{\textsc{ControlTac}\xspace}

\author{
Dongyu Luo$^{*\mkern3mu\dagger}$\kern0.18em,
Kelin Yu$^{*}$\kern0.18em, Amir-Hossein Shahidzadeh,\\
\textbf{Cornelia Fermuller, Yiannis Aloimonos, Ruohan Gao}\\[0.8em]
University of Maryland, College Park%
{\thanks{Indicates equal contribution.}}%
{\thanks{Dongyu Luo is affiliated with The University of Hong Kong. The work was done during an internship at the University of Maryland.}}
}

\begin{document}

\title{\name: Scaling Tactile Data with \\Physically Controlled Tactile Image Generation} 

\maketitle

\vspace{-2pt}
\begin{abstract}
  Vision-based tactile sensing is widely used in perception, reconstruction, and robotic manipulation, yet collecting large-scale tactile data remains costly due to diverse sensor-object interactions and inconsistencies across sensor instances. Existing approaches to scaling tactile data---simulation and free-form tactile generation---often yield unrealistically rendered signals with poor transfer to highly dynamic real-world tasks. We propose \name, a two-stage controllable tactile image generation framework that generates realistic tactile images conditioned on a single reference tactile image, contact force, and contact pose. By grounding generation in these important physical priors, \name synthesizes realistic samples across different sensors while effectively capturing task-relevant variations. Across a series of downstream tasks and real-world experiments, such as object insertion, imitation learning, and object weighting, the augmented datasets using our approach consistently improve performance and demonstrate practical utility in dynamic real-world settings. Project page: \url{https://dongyuluo.github.io/controltac/}.
\end{abstract}


\begin{figure*}[h!]
    \centering
    \vspace{-12pt}
    \includegraphics[width=1\textwidth]{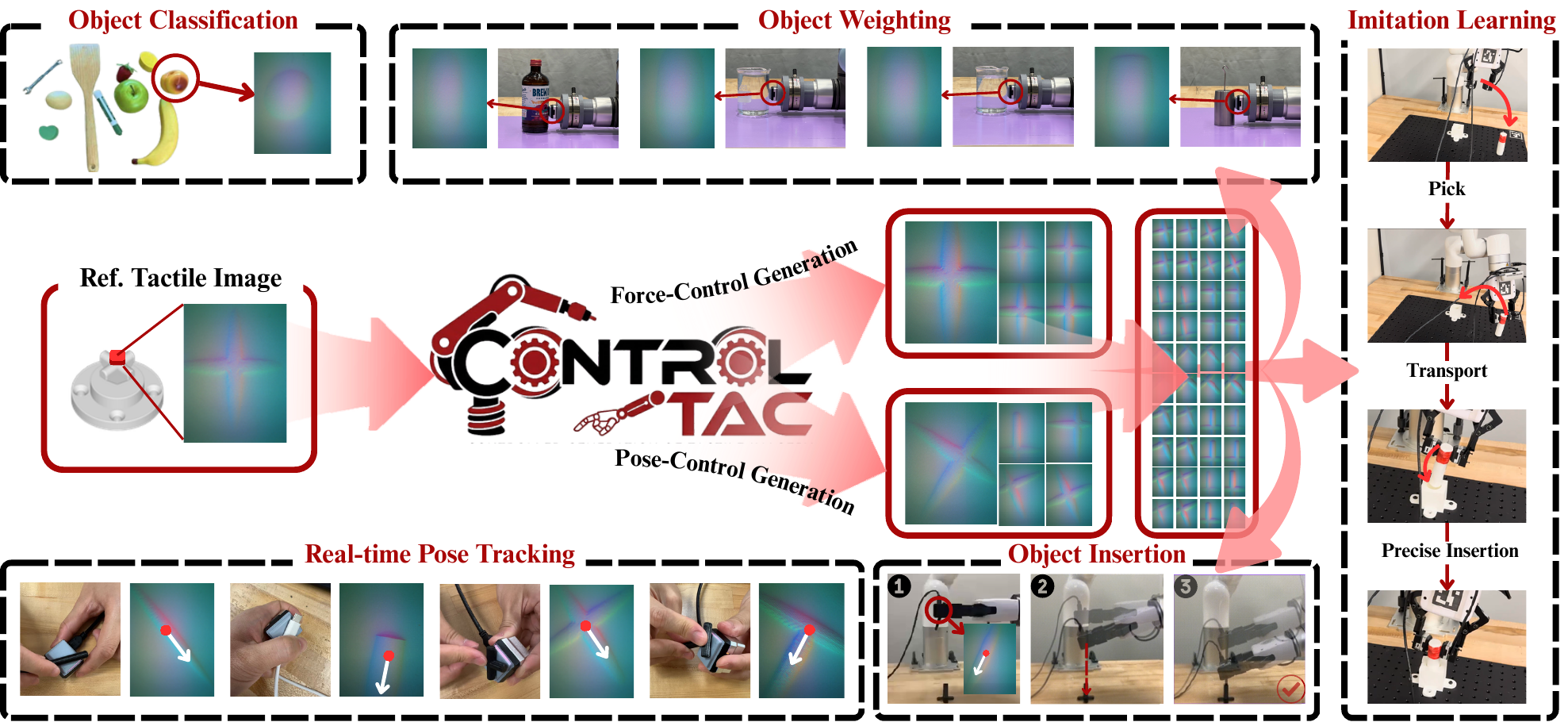}
    \caption{\textbf{Overview of \name.} Given a single reference image, \name performs force- and pose-conditioned generation to synthesize millions of realistic tactile images (center). This augmented dataset enhances various downstream applications, including object classification, weight estimation, real-time pose tracking, object insertion, and training imitation learning policies.}
\label{fig:overview}
\end{figure*}


\section{Introduction}


Tactile sensing provides critical physical interaction feedback that vision alone cannot capture, making it indispensable for advanced perception. Among various modalities, vision-based tactile sensors~\cite{tactileoverview, gelsight} are widely used to capture high-resolution elastomeric deformations as contact images for material classification~\cite{texture}, 3D reconstruction~\cite{normalflow}, and robotic manipulation~\cite{tactilerl}. However, unlike object images that can be easily collected or crawled from the internet, acquiring large-scale tactile datasets remains costly, as tactile sensing inherently requires physical contact across a wide range of poses and forces. Moreover, tactile images exhibit substantial variation due to differences in gel properties and sensor lighting conditions, which limits cross-sample transfer and reuse. These factors make scaling tactile data significantly more challenging than scaling vision data, underscoring the need for effective tactile-specific data augmentation strategies.

To scale tactile datasets via augmentation, existing approaches mainly follow two paradigms: tactile simulation~\cite{tacto, taxim} and cross-modal generation (\textsc{Vis2Tac}/\textsc{Text2Tac})~\cite{texttoucher, touchandgo}. However, tactile simulators suffer from a substantial sim-to-real gap due to imperfect soft-body modeling and optical inconsistencies, while cross-modal generative methods lack the controllability and fidelity needed for challenging dynamic manipulation scenarios. To leverage generative models while addressing these limitations, our key insight is that tactile synthesis should be conditioned on important physically meaningful parameters—contact forces and contact poses. Moreover, rather than generating tactile images from scratch, using a single real tactile image as a reference provides an efficient anchor for fidelity, preserving essential details in geometry, texture, and sensor illumination.

\begin{figure*}[t]
    \centering
    \begin{minipage}{0.5\textwidth}
        \centering
        \scriptsize
        \begin{tabular}{@{}l|ccc@{}}
        \toprule
         & Realism & Variation & Controllable \\
        \midrule
        \textsc{Text2Tac}       & Low    & Low    & \ding{55} \\
        \textsc{Vis2Tac}     & Low    & Medium & \ding{55}\\
        \textsc{Simulation}  & Medium & Medium & \ding{51} \\
        \name  & High   & High   & \ding{51} \\
        \bottomrule
        \end{tabular}
        \label{tab:representations}
    \end{minipage}%
    \begin{minipage}{0.50\textwidth}
        \centering
        \includegraphics[width=1\textwidth]{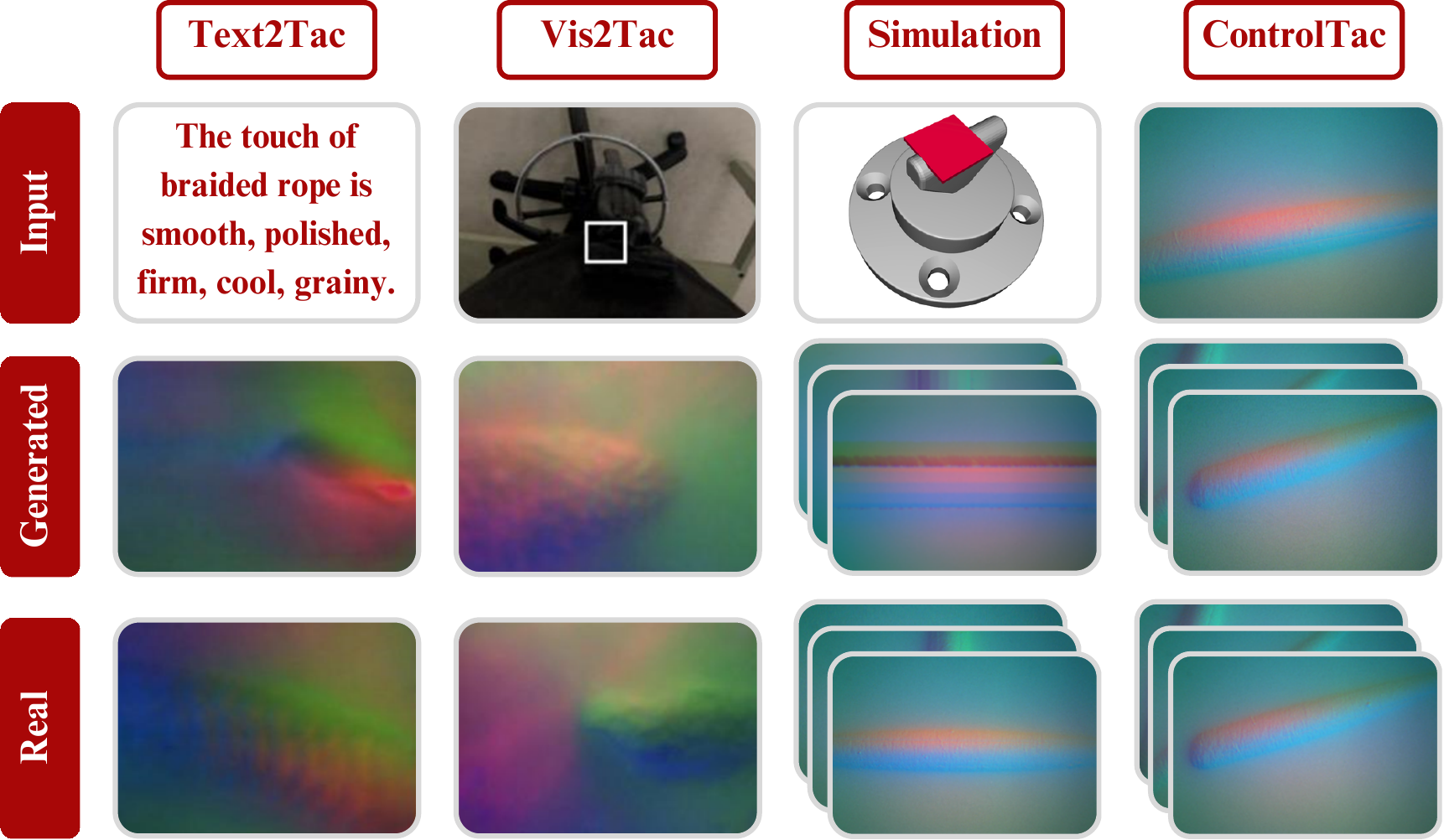}
    \end{minipage}
    \caption{
    \textbf{Comparison of tactile data augmentation approaches.} We compare \name with three three paradigms for tactile data augmentation---\textsc{Text2Tac}~\cite{texttoucher, binding}, \textsc{Vis2Tac}~\cite{visgel, tarf, touchinganerf}, and \textsc{Simulation}~\cite{taxim, difftactile, taccel}---along three criteria: \texttt{(1)} whether the method produces visually realistic tactile images, \texttt{(2)}  whether it can generates diverse outputs from a single input (rather than collapsing to a mean image), and \texttt{(3)} whether it supports controllability via physical inputs.
    }
    \label{fig:tac_gen}
\end{figure*}

To realize this intuition, we propose \name, a two-stage controllable tactile image generation framework that explicitly injects physical realism and contact dynamics into tactile data augmentation. First, given a single reference tactile image, \name conditions on relative 3D force vectors to synthesize target images with accurate elastomer deformations under varying forces. Second, a pose-guided adapter takes a 2D contact mask as input to precisely control contact locations across the sensor surface. By jointly modeling force dynamics and pose diversity, \name generates physically plausible and diverse tactile data from a single reference image, substantially outperforming existing augmentation methods (Fig.~\ref{fig:tac_gen}). Furthermore, \name can achieve cross-sensor generation, enabling data reuse across different sensor instances by transferring tactile features across distinct gel and lighting conditions.

Extensive experiments show that \name yields consistent improvements across downstream tasks including object classification, force estimation, and pose estimation. We demonstrate its effectiveness in challenging real-world manipulation settings, including object weighting, real-time pose tracking, and precise peg insertion. Additionally, we apply \name to perform tactile data augmentation on teleoperated robot datasets, where the augmented data significantly improves imitation learning performance. In summary, our contributions are threefold:  \texttt{(1)} we propose physically controlled tactile image generation as an approach for scaling tactile data via augmentation;  \texttt{(2)} we introduce \name, a framework leveraging a single reference image to enable explicit control over dynamic forces, diverse poses, and cross-sensor transfer; and  \texttt{(3)} we validate \name across broad downstream applications, real-world manipulation, and a challenging contact-rich Peg-In-Hole imitation learning task trained with augmented dataset.
  
\section{Related Work}

\textbf{Vision-based Tactile Datasets.}\quad Vision-based tactile sensors~\cite{gelsight, gelslim, digit, 9dtact} have gained momentum due to their high spatial resolution and rich appearance cues, and are widely applied to perception and manipulation tasks~(cf. Appendix~\ref{app:tactile_details}). To address the scarcity of tactile data, many prior works focus on collecting large-scale real-world datasets~\cite{touchandgo, tarf, objectforlder, visgel}. However, the data quantity remains limited, and variability across sensors makes reuse for downstream robotics tasks difficult. Another approach is simulation~\cite{tacto, taxim, difftactile, objectfolder2, pbr}, widely adopted for pre-training~\cite{t3, sparsh, sensorinv, anytouch} and Sim2Real transfer~\cite{braille, objectfolder2, difftactile}. Yet, bridging the Sim2Real gap remains a major challenge, as high-quality transfer still requires large real datasets for co-training~\cite{tacto} or generative models for domain adaptation~\cite{braille}. Differently, we introduce \name, a controllable tactile generation framework to scale up tactile data under different physical conditions.


\textbf{Tactile Image Generation.}\quad Text-to-tactile generation~\cite{texttoucher, binding}, vision-to-tactile generation~\cite{visgel, tarf, touchandgo, touchinganerf, generating}, and cross-sensor generation~\cite{touch2touch} have been explored for representation learning~\cite{anytouch, t3}, contact localization~\cite{tarf, obj_folder_bench}, classification~\cite{tarf, touchinganerf}, and retrieval~\cite{obj_folder_bench, binding} tasks. However, free-form generation from visual images often produces low-quality outputs, and only a single tactile image can be generated from one contact location, limiting coverage of dynamic variations in downstream tasks. To address this, we propose a conditional diffusion model that generates tactile images for data augmentation guided by physical constraints and priors.


 \textbf{Conditional Image Generation.}\quad Conditional image generation aims to generate images guided by structured inputs such as class labels, text, or physical parameters. Early methods based on conditional GANs~\cite{CGAN1, CGAN2, CGAN3, CVAE1, CVAE2, CVAE3, CGAN, CVAE} demonstrate the feasibility of conditional generation but often suffer from limitations in image quality and training stability~\cite{GanL1,GanL2,GanL3,GanL4}. More recently, diffusion models~\cite{ddpm, ddim, score, stablediffusion, pixart, sana} achieve high-quality image generation through gradual denoising, while ControlNet~\cite{ControlNet, controllable} introduces auxiliary networks to inject structural conditions such as edge maps, depth maps, or human poses into the generation process. Inspired by, but distinct from the prior work above, we tackle the new problem of controllable tactile image generation.

\section{Methodology}\label{sec:method}
\begin{figure}[t]
    \centering
    \includegraphics[width=1\textwidth]{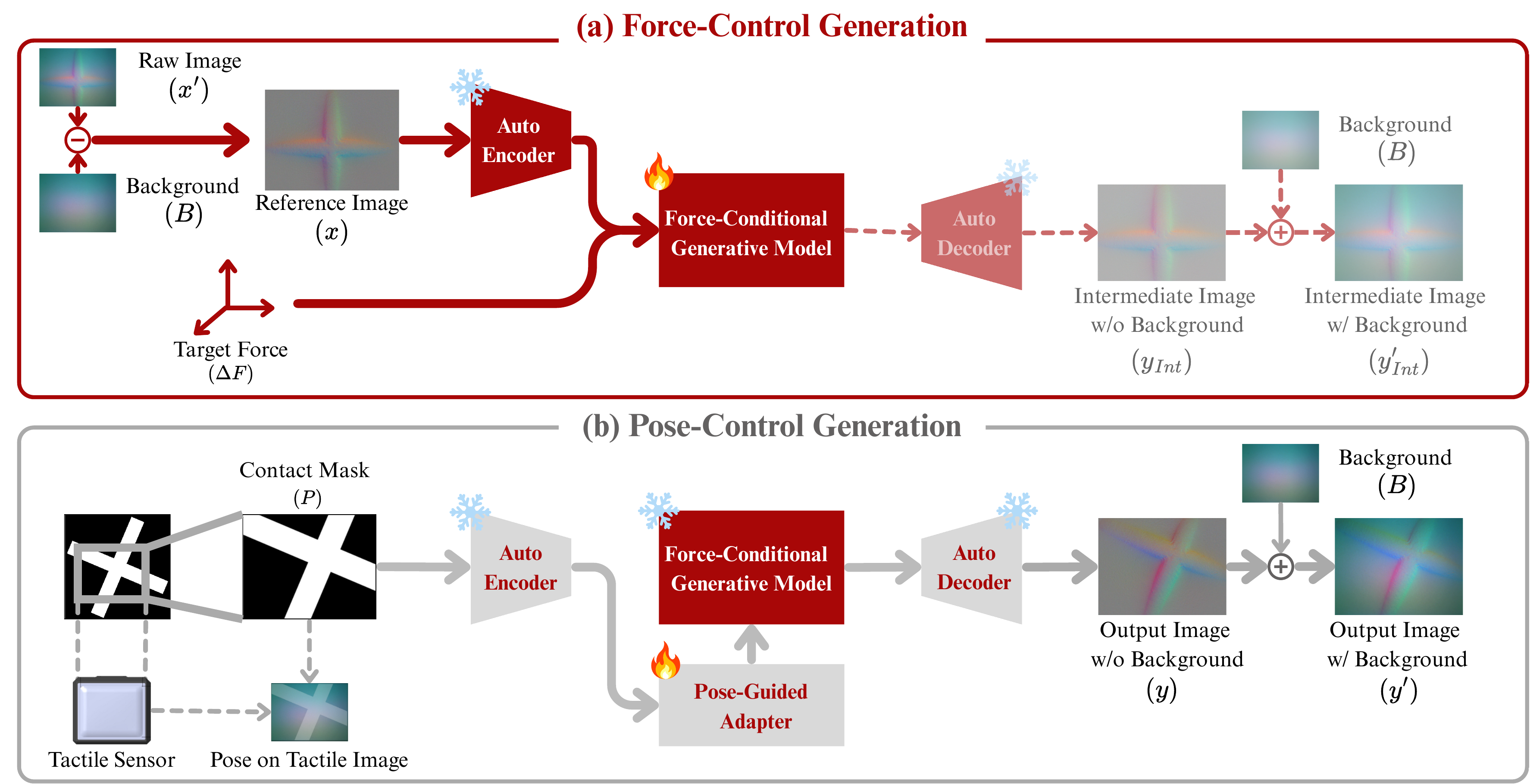}
    \caption{\textbf{Two-stage \name framework.} 
    \textbf{(a) Force-Control:} The raw image $x'$ is background-subtracted and encoded into latent space. Conditioned on the target force $\Delta F$, the generator synthesizes a force-adjusted intermediate image $y_{Int}$. 
    \textbf{(b) Pose-Control:} An object-specific contact mask $P$ is transformed via rigid 2D operations to serve as an explicit spatial constraint, guiding the model to generate the final tactile image $y$ that satisfies both $\Delta F$ and the target pose.}
\label{fig:framework}
\vspace{-0.5cm}
\end{figure}

We present \name, a controllable framework that augments tactile data from a single reference image conditioned on contact forces and poses. To ensure realistic generation, \name preserves contact geometry and color using the reference image, while modeling physical variations via a two-stage conditional pipeline. Below, we first review vision-based tactile sensing (Sec.~\ref{sec:background}), and then introduce our framework which incorporates force and pose as physical priors (Sec.~\ref{sec:controlnet}). Finally, we demonstrate \name's application to downstream object classification, force estimation, and pose estimation (Sec.~\ref{sec:augment}). The overall framework is illustrated in Fig.~\ref{fig:framework}.

\subsection{Background on Vision-based Tactile Sensing}
\label{sec:background}
Vision-based tactile sensors capture elastomeric gel deformations via onboard cameras, but variations in gel properties and lighting lead to substantial appearance discrepancies~\cite{gelsight, digit, gelslim, t3, sensorinv}. Accurate 3D contact force estimation from these observations is highly non-trivial, as the physical mappings are heavily entangled with material properties, local geometries, and contact patches~\cite{ifem}. Consequently, comprehensively scaling real-world visuo-tactile data across diverse forces, poses, and sensor variations is prohibitively expensive and largely impractical~\cite{feelanyforce, 9dtact}. To address this data bottleneck, we propose a controllable generative framework to synthesize physically realistic tactile images at scale, evaluating our approach on the widely-used GelSight Mini sensor~\cite{mimictouch, anytouch, t3}.

\subsection{Two-stage Conditional Tactile Generation Framework}
\label{sec:controlnet}
To model real-world interaction variance, we propose a two-stage conditional generation framework that decouples physical priors into contact force and spatial pose. Simultaneous generation often entangles representations, causing force changes to alter contact geometry. Thus, our pipeline operates sequentially: \texttt{(1)} a force-control stage modifying reference image deformation and shading for a target force, and \texttt{(2)} a pose-control stage structurally guiding generation to a target pose.
\textbf{Stage 1: Force-Control Generation.}\quad This stage introduces realistic force-induced deformations while preserving textures and contact regions. We condition the model on a reference image (initial contact) to implicitly capture geometry, texture, color, and gel elasticity. To mitigate cross-sensor domain gaps (e.g., lighting), the background $B$ is subtracted from the raw image $x'$ to isolate contact features in $x$. For efficiency, $x$ is encoded into a compressed latent space via a pre-trained autoencoder. Finally, a conditional diffusion-based generator takes this latent representation and a target relative force vector $\Delta F$ to synthesize a force-adjusted latent feature, which can be decoded into an intermediate tactile image $y_{Int}$.

\textbf{Stage 2: Pose-Control Generation.}\quad This stage uses a compact, object-specific binary contact mask $P$ as a global pose-control signal to augment data across diverse contact poses. Manually aligned with the reference image, $P$ approximates the global contact surface, which can extend beyond the sensor area. Crucially, $P$'s shape and scale remain fixed; contact pose variations are modeled solely via 2D rigid transformations on the mask. Masks are aligned with ground-truth tactile images with a precision of $\pm3$ pixels and $\pm1^\circ$ during training, and can be transformed to arbitrary target poses during inference. This design avoids two pitfalls: \texttt{(1)} center-angle representations $(x, y, \theta)$ fail when objects exceed sensor size, and \texttt{(2)} edge detection (e.g., Canny~\cite{canny}) yields force-inconsistent boundaries. Thus, $P$ serves as a fixed template decoupled from force variations. Inspired by ControlNet~\cite{ControlNet}, we freeze the pretrained force-conditional backbone and introduce a pose-guided adapter. Taking mask $P$ as a spatial constraint, the adapter guides generation to the target contact pose while preserving the learned force-deformation mapping. Consequently, the final tactile image $y$ accurately conforms to both target force $\Delta F$ and configuration $P$.

\subsection{Downstream Applications Leveraging \name}
\label{sec:augment}
\name is a versatile, scalable framework for data-hungry tactile perception tasks. By varying force and pose parameters, it synthesizes large, diverse datasets from minimal real-world data. To evaluate data realism and utility, we benchmark on three representative downstream tasks: a discrete categorization task (object classification), a continuous regression task highly sensitive to physical dynamics (force estimation), and a spatial reasoning task (contact pose estimation).

\textbf{Object Classification.}\quad Following standard practices in \textsc{Vis2Tac}~\cite{tarf} and \textsc{Text2Tac}~\cite{texttoucher}, this task requires distinguishing unseen objects under varying contact states to validate the quality of our generated data. To ensure a comprehensive assessment, we report classification results across multiple architecture-agnostic models (details in App.~\ref{app:classifier})..

\textbf{Force Estimation.}\quad 3D contact force estimation is a sensitive regression task where minor deformation errors severely degrade accuracy~\cite{anytouch2}, rendering it ideal for evaluating realism. Following~\cite{feelanyforce}, we train a force estimator to predict 3D force vectors (normal and 2D shear) from raw tactile inputs. We investigate whether augmenting limited real data with our generated samples can approach the oracle performance of the model trained on the full real dataset—a capability undemonstrated by prior augmentation methods (see App.~\ref{app:force_estimator} for implementation details).

\textbf{Pose Estimation.}\quad Following the force estimation setup, we train a pose estimator to evaluate the spatial precision and diversity of our generated samples. Specifically, we assess whether a pose estimator trained on data generated by \name can outperform one trained on the complete real dataset in predicting 2D contact locations $(x, y)$ and sensor rotation angles (details in App.~\ref{app:pose_estimator}).%

\section{Experiments}\label{sec:exp}
We evaluate \name on tactile image generation quality (Sec.~\ref{sec:gen_qual}), downstream data augmentation tasks (Sec.~\ref{sec:downstream}), and real-world robotic deployment (Sec.~\ref{sec:real}).

\textbf{Baselines.}\quad We evaluate \name against four baseline categories. For tactile \textsc{Simulator}, we implement a sampling-based simulation baseline using Taxim~\cite{taxim} (with manual pose control and $z$-displacement to normal force mapping), representing methods that leverage reference images to align with real data~\cite{taxim, difftactile, taccel, tacex}. For cross-modal generation, \textsc{Text2Tac}~\cite{texttoucher} is used in an object classification case study to highlight the limitations of free-form generation~\cite{binding, visgel, tarf, touchinganerf}. For \textsc{Real2Sim}, we adapt the generative approach from~\cite{sim2real} to synthesize tactile images from simulated ones. Finally, we compare \name against two internal ablations: \texttt{(1)} \textsc{Hybrid} (joint force-pose diffusion conditioning) and \texttt{(2)} \textsc{Separate}-Control (sequential pose generator training on force-model images). Implementation, architecture, and deployment details are in App.~\ref{app:implementation}, including precise insertion task and contact-rich imitation learning.

\subsection{Evaluating Tactile Image Generation Quality}
\label{sec:gen_qual}
\begin{figure}[t!]
    \centering
    \includegraphics[width=1\textwidth]{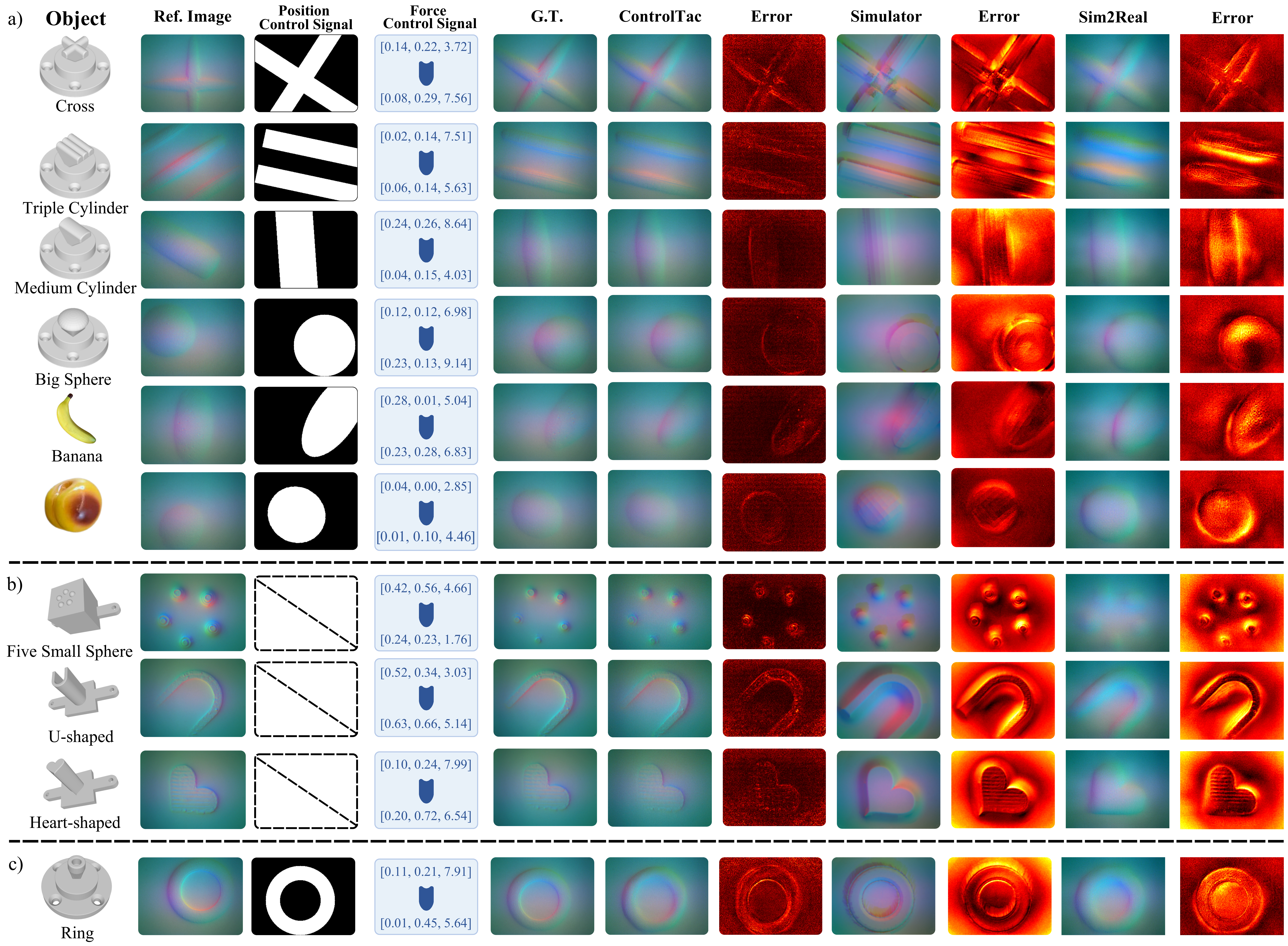}
    \caption{\textbf{Qualitative generation results across diverse objects and contact conditions.} Visual comparison of our \name against baselines on: \textbf{(a)} seen (rows 1--4) and unseen (rows 5--6) objects from FeelAnyForce~\cite{feelanyforce}, \textbf{(b)} unseen objects from AnyTouch2~\cite{anytouch2}, and \textbf{(c)} a failure case. See App.~\ref{app:full_vis} for the complete set of qualitative results and detailed error map breakdowns.}
    \label{fig:vis_all}
\end{figure}

Table~\ref{tab:method_comparison} and Fig.~\ref{fig:vis_all} report evaluation results across twelve seen and unseen objects (App.~\ref{app:implementation}). On AnyTouch2, we only evaluate force-controlled generation due to the lack of precise contact masks. \name outperforms \textsc{Simulator}~\cite{taxim, difftactile, taccel} and \textsc{Sim2Real} baselines in SSIM and MSE. Qualitative results (Fig.~\ref{fig:vis_all}) show that \name captures complex force-induced deformations and fine textures, even on unseen objects. Furthermore, \name surpasses all internal ablations (App.~\ref{app:full_vis}), validating our two-stage framework.

\begin{figure}[htpb]
    \centering
    \begin{subfigure}[t]{0.48\textwidth}
        \centering
        \vspace{0pt}
        \includegraphics[width=\textwidth]{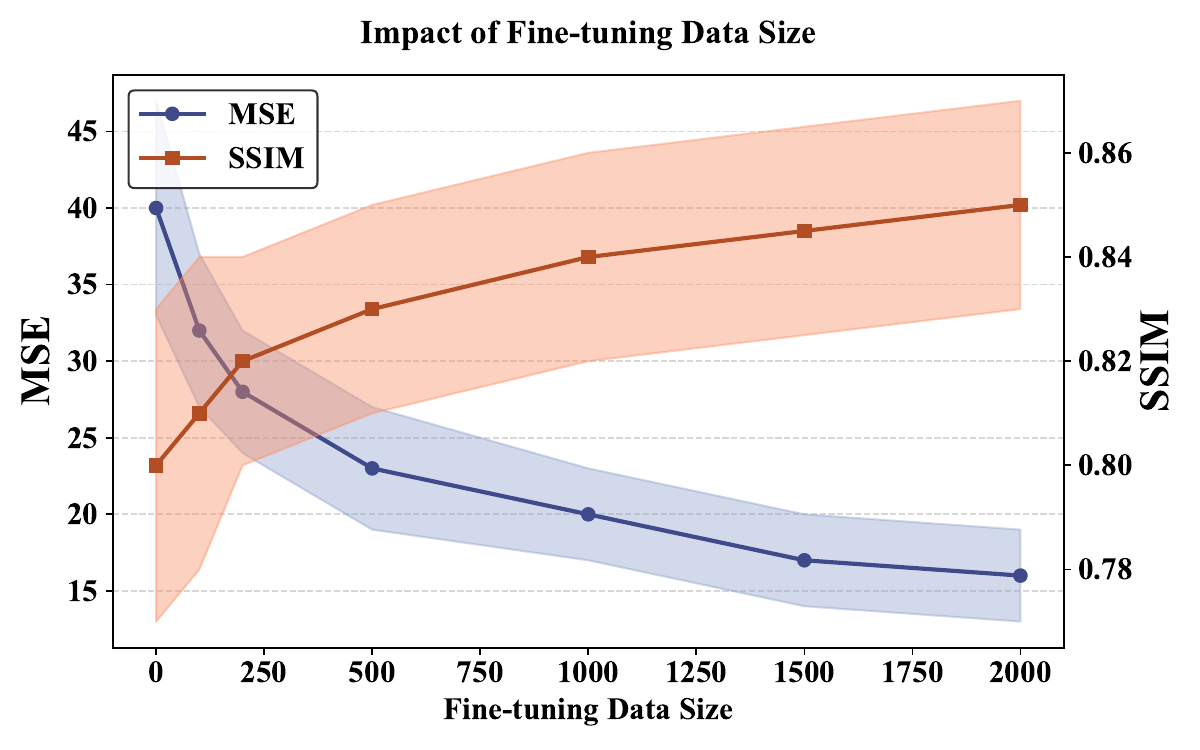}
    \end{subfigure}
    \begin{subfigure}[t]{0.48\textwidth}
        \centering
        \vspace{0pt}
        \includegraphics[width=1\textwidth]{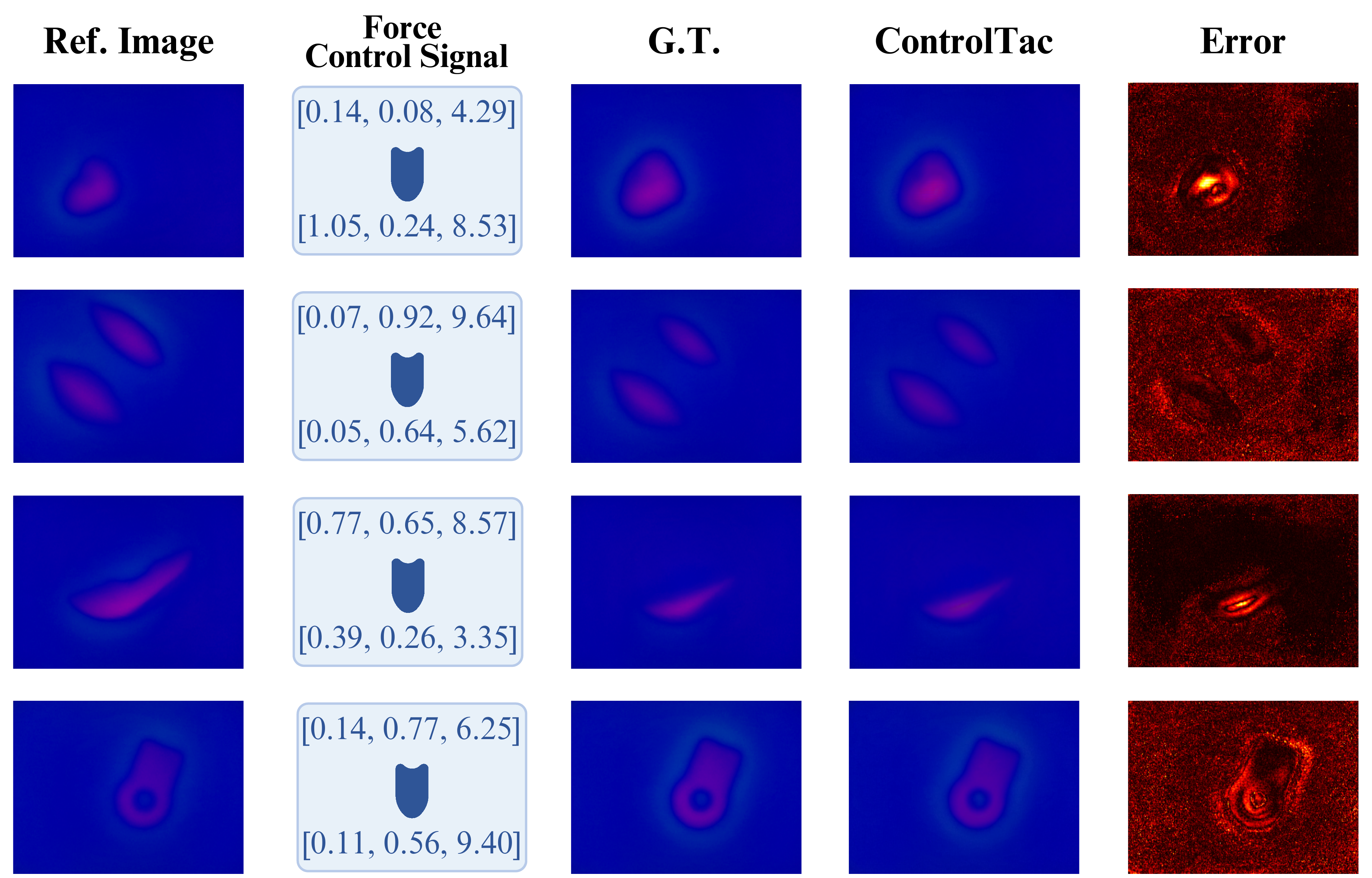}
    \end{subfigure}
    \caption{\textbf{Evaluation of cross-sensor generation capabilities on the 9DTact dataset.} \textbf{(a)} Impact of fine-tuning data size on MSE$\downarrow$ and SSIM$\uparrow$ metrics (the shaded area denotes the standard deviation). \textbf{(b)} Visualization of generated results across different sensors.}
    \label{fig:cross_sensor_eval}
    \vspace{-0.4cm}
\end{figure}

\begin{table}[t!]
\centering
\footnotesize
\renewcommand{\arraystretch}{1.2} 
\caption{\textbf{Quantitative Results.} Comparison of MSE$\downarrow$ and SSIM$\uparrow$ (mean $\pm$ SD) across two datasets. \textsc{Hybrid}, \textsc{Separate}, and \textsc{Simulator} denote the Hybrid Force-Pose Model, Separate-Control Pipeline, and simulation baselines~\cite{taxim, difftactile, taccel}, respectively. On AnyTouch2~\cite{anytouch2}, the two ablations cannot be evaluated due to the lack of accurate contact masks.}
\vspace{1pt}
\resizebox{\textwidth}{!}{%
\begin{tabular}{l 
>{\centering\arraybackslash}p{2.2cm} >{\centering\arraybackslash}p{2.2cm} 
>{\centering\arraybackslash}p{2.2cm} >{\centering\arraybackslash}p{2.2cm} 
>{\centering\arraybackslash}p{2.2cm} >{\centering\arraybackslash}p{2.2cm}}
\toprule
\multirow{2}{*}{Method} & \multicolumn{2}{c}{Seen Objects} & \multicolumn{2}{c}{Unseen: FeelAnyForce} & \multicolumn{2}{c}{Unseen: AnyTouch2} \\
\cmidrule(lr){2-3} \cmidrule(lr){4-5} \cmidrule(lr){6-7}
 & MSE$\downarrow$ & SSIM$\uparrow$ & MSE$\downarrow$ & SSIM$\uparrow$ & MSE$\downarrow$ & SSIM$\uparrow$ \\
\midrule
\textsc{Simulator}~\cite{taxim, difftactile, taccel} & 1054 ± 19 & 0.68 ± 0.03 & 1065 ± 23 & 0.69 ± 0.03 & 2157 ± 14 & 0.61 ± 0.04 \\
\textsc{Sim2Real}~\cite{sim2real} & 239 ± 17 & 0.74 ± 0.03 & 253 ± 25 & 0.73 ± 0.03 & 545 ± 31 & 0.70 ± 0.05 \\
\textsc{Hybrid} & 31 ± 5 & 0.81 ± 0.04 & 37 ± 6 & 0.75 ± 0.04 & - & - \\
\textsc{Separate} & 157 ± 8 & 0.79 ± 0.04 & 199 ± 11 & 0.72 ± 0.05 & - & - \\
\textbf{Ours} & \textbf{23 ± 2} & \textbf{0.83 ± 0.03} & \textbf{26 ± 3} & \textbf{0.79 ± 0.04} & \textbf{29 ± 2} & \textbf{0.81 ± 0.02} \\
\bottomrule
\end{tabular}%
}
\label{tab:method_comparison}
\end{table}
For cross-sensor evaluation on the 9DTact~\cite{9dtact} dataset, \name demonstrates strong zero-shot generalization and achieves comparable performance to the training sensor with fine-tuning on small dataset with only $2000$ images (Fig.~\ref{fig:cross_sensor_eval}(a)). Visualizations in Fig.~\ref{fig:cross_sensor_eval}(b) further confirm its capability to capture unique lighting and texture details of unseen sensors.

We provide robustness analyses on mask perturbations (App.~\ref{app:mask}) and discuss failure cases (App.~\ref{app:failure}). App.~\ref{app:force_only} visualizes force-controlled generation alone. Additionally, depth visualizations (App.~\ref{app:depth}) demonstrate that \name captures true surface geometry rather than overfitting to brightness.

\subsection{Results on Downstream Applications}\label{sec:downstream}
\name can benefit a wide variety of robotic perception and manipulation applications by providing abundant, high-fidelity tactile data. We demonstrate this by evaluating our generated data across three tasks, scaled from fundamental classification to challenging physical state estimation.

\begin{figure}[htpb]
    \centering
    \begin{minipage}{0.48\textwidth}
        \centering
        \scriptsize 
        \setlength{\tabcolsep}{5pt}
        \begin{tabular}{l|ccc}
        \toprule
        \textbf{Method} & \textbf{CNN} & \textbf{\makecell[c]{ViT\\(Scratch)}} & \textbf{\makecell[c]{ViT\\(ImageNet)}} \\
        \midrule
        Geo+Col~\cite{data_aug1} & 0.69 & 0.65 & 0.79 \\
        \textsc{Simulator}~\cite{taxim,difftactile,taccel} & 0.79 & 0.85 & 0.92 \\
        \textbf{Ours} & \textbf{0.87} & \textbf{0.95} & \textbf{0.99} \\
        \midrule
        \textsc{Text2Tac}*~\cite{texttoucher} & 0.22 & 0.16 & 0.14 \\
        \bottomrule
        \end{tabular}
        \label{tab:classification} 
    \end{minipage}
    \hfill
    \begin{minipage}{0.48\textwidth}
        \centering
        \includegraphics[width=1\textwidth]{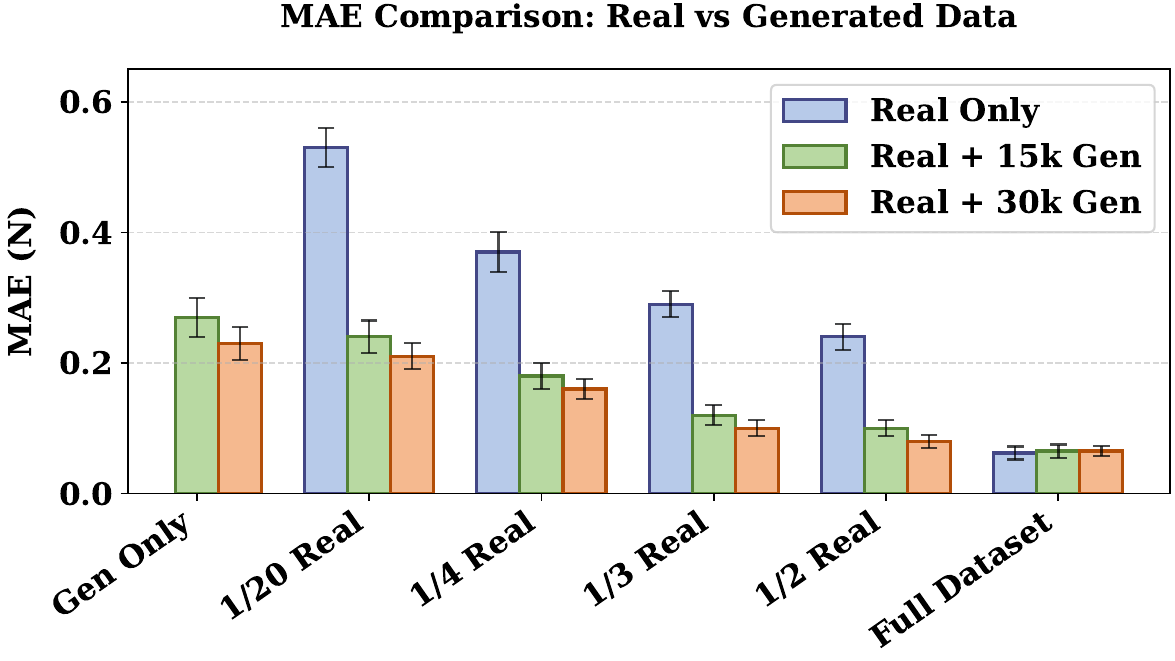} 
    \end{minipage}
    
    \caption{
    \textbf{(a)} Accuracy$\uparrow$ comparisons on object classification.
    \textbf{(b)} Impact of data scaling on force estimation. Comparison of mean MAE$\downarrow$ ($\pm$ SD, $N=1200$) across different scales of real and augmented (15k, 30k) training datasets.
    }
    \label{fig:combined_results}
    \vspace{-0.3cm}
\end{figure}

\textbf{Object Classification.}\quad To evaluate generalizability, we perform unseen object classification—the only downstream task compatible with \textsc{Vis2Tac} and \textsc{Text2Tac} due to their lack of control. Using one reference image for six unseen objects, we generate tactile data under varying forces and poses. We compare \name against geometric/color augmentations~\cite{data_aug1,data_aug2} and a simulator baseline. Fig.~\ref{fig:combined_results}(a) shows that \name consistently outperforms all baselines. \textsc{Text2Tac} performs poorly as free-form generation lacks physical grounding, and the simulator is limited by the sim-to-real gap. Details are in App.~\ref{app:text2tac_class} and App.~\ref{app:class}.

\textbf{Force Estimation.}\quad Given GelSight's high sensitivity to subtle force variations~\cite{gelsight, feelanyforce}, we evaluate \name's capability in generating realistic tactile images across diverse forces ($1$--$10\,\text{N}$, $0.1\,\text{N}$ precision) and poses. We synthesize 15,000 to 30,000 images to co-train a force estimator with varying subsets of real data. As shown in Fig.~\ref{fig:combined_results}(b), \name effectively covers pose and force variations, reducing Mean Absolute Error (MAE) when real data is scarce. Notably, supplementing only one-third of the real dataset with \name data matches the full-dataset performance, whereas using that real subset alone yields poor results. While combining all real and synthetic data slightly underperforms the full real dataset alone—as the complete FeelAnyForce dataset already achieves near-oracle performance—collecting such extensive real data remains impractical. Scaling laws, force generation, and sensitivity analyses are detailed in App.~\ref{app:force}, \ref{app:full_comparison}, and \ref{app:force_sensitivity}.

\textbf{Pose Estimation.}\quad
Unlike geometry-dependent 3D estimators~\cite{tac2pose}, 2D pose estimation generalizes better to unknown objects and diverse tasks (e.g., cable insertion~\cite{cable}, active perception~\cite{tactilefilter}). However, while PCA fails on asymmetric shapes, flexible learning-based estimators~\cite{tacto} demand extensive labeled data. To bridge this gap, we adopt the learning-based paradigm but automate data generation via \name, which extracts pose labels from 2D contact masks during image synthesis.

\begin{table}[t!]
\centering
\footnotesize
\renewcommand{\arraystretch}{1.2} %
\caption{\textbf{Contact pose estimation errors} (X/Y in mm, Angle in degrees) $\downarrow$ for seen and unseen objects (mean $\pm$ standard deviation). We evaluate performance on two seen (\textit{Cylinder}, \textit{Cross}) and two unseen (\textit{T-shape}, \textit{USB}) objects. ``Fixed'' and ``unfixed'' denote training with a constant (6.5N) or varying contact force, respectively. Bold numbers indicate the best performance.}
\vspace{1pt}
\label{tab:position-estimation}
\resizebox{1\textwidth}{!}{%
\begin{tabular}{l|>{\centering\arraybackslash}p{1.5cm}>{\centering\arraybackslash}p{1.5cm}>{\centering\arraybackslash}p{1.5cm}|>{\centering\arraybackslash}p{1.5cm}>{\centering\arraybackslash}p{1.5cm}>{\centering\arraybackslash}p{1.5cm}}
\toprule
\multirow{2}{*}{\textbf{Seen Objects}}
& \multicolumn{3}{c}{\textit{Cylinder (3 Types)}} 
& \multicolumn{3}{c}{\textit{Cross}} \\
\cmidrule(lr){2-4} \cmidrule(lr){5-7}
& X & Y & Angle & X & Y & Angle \\
\midrule
PCA~\cite{cable} 
& 15 ± 6 & 13 ± 3 & 22 ± 4 
& 56 ± 9 & 19 ± 5 & 18 ± 5 \\
Real 
& 8 ± 3 & 8 ± 2 & 4 ± 2 
& 6 ± 2 & 6 ± 2 & 2 ± 2 \\
\textsc{Simulator}~\cite{taxim, difftactile, taccel} 
& 17 ± 3 & 15 ± 3 & 6 ± 2 
& 19 ± 3 & 18 ± 2 & 5 ± 2 \\
\textsc{Simulator} + Real 
& 12 ± 3 & 13 ± 3 & 6 ± 3 
& 17 ± 4 & 16 ± 3 & 4 ± 3 \\
Ours (fixed) 
& 9 ± 2 & 8 ± 2 & 5 ± 1 
& 7 ± 2 & 9 ± 2 & 4 ± 2 \\
\textbf{Ours (unfixed)} 
& \textbf{4 ± 1} & \textbf{5 ± 1} & \textbf{3 ± 1} 
& \textbf{3 ± 1} & \textbf{4 ± 1} & \textbf{1 ± 1} \\
\textbf{Ours (unfixed) + Real} 
& \textbf{3 ± 1} & \textbf{4 ± 1} & \textbf{3 ± 1} 
& \textbf{2 ± 2} & \textbf{4 ± 1} & \textbf{1 ± 1} \\
\midrule
\multirow{2}{*}{\textbf{Unseen Objects}}
& \multicolumn{3}{c}{\textit{T-shape}} 
& \multicolumn{3}{c}{\textit{USB}} \\
\cmidrule(lr){2-4} \cmidrule(lr){5-7}
& X & Y & Angle & X & Y & Angle \\
\midrule
\textbf{Ours (unfixed)} 
& \textbf{4 ± 2} & \textbf{5 ± 1} & \textbf{2 ± 1} 
& \textbf{5 ± 1} & \textbf{4 ± 1} & \textbf{3 ± 1} \\
\bottomrule
\end{tabular}%
}
\end{table}

To evaluate the robustness and generalization of \name, we train estimators on four known objects (a cross and three cylinders) and two unseen shapes (a T-shape and a Type-C connector), testing the latter on a new sensor instance to assess hardware transferability (App.~\ref{app:sensor}). The test sets comprise 30 annotated poses per object under varied forces. As shown in Table~\ref{tab:position-estimation}, estimators trained solely on \name-generated data achieve strong performance across all objects, outperforming models trained on real data alone. This is because \name's diverse synthesis effortlessly covers physical gaps in contact poses and dynamics (App.~\ref{app:special}). Furthermore, \name outperforms realism-lacking simulators~\cite{taxim, difftactile, taccel} and traditional PCA baselines. Finally, training with varying forces rather than a fixed force ($6.5\,\text{N}$) significantly improves robustness to force variations. (Extended results and visualizations are in App.~\ref{app:push} and \ref{app:vis_taxim}.)

\subsection{Real-World Experiments}
\label{sec:real}
In this section, we deploy the estimators and policies trained on our augmented dataset across four dynamic real-world tasks to demonstrate their robust generalization, millimeter-level precision, and effectiveness in downstream manipulation.

\begin{figure}[htpb]
\centering
\includegraphics[width=1\textwidth]{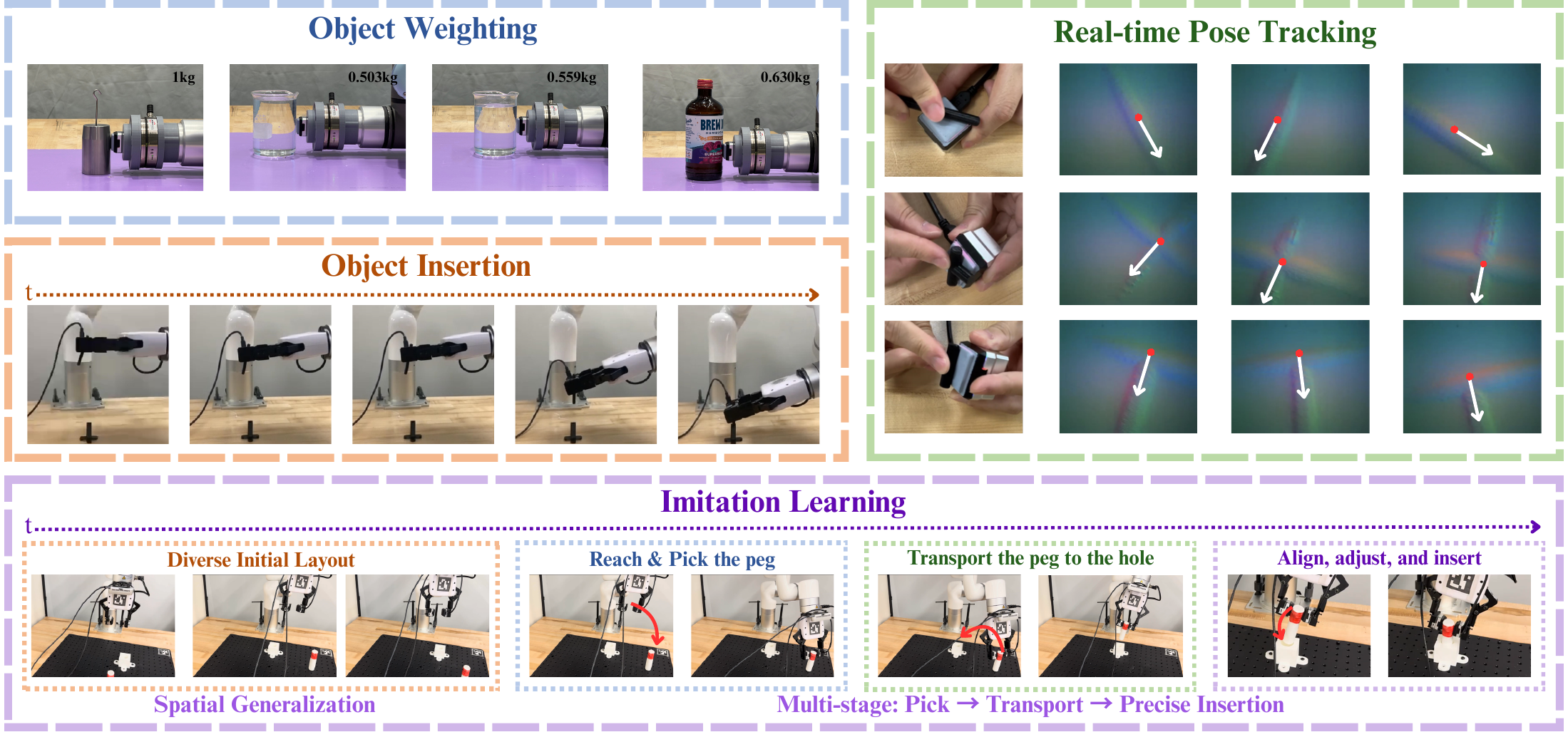}
\caption{\textbf{Real-World Applications.} Real-world experiments demonstrate estimators or policies trained with \name-generated data on object weighting, real-time pose tracking, object insertion, and imitation learning.
}
\label{fig:real_world}
\end{figure}

\textbf{Object Weighting.}\quad We estimate the pushing force between a UR5 robot and four objects (metal weight, water cylinders, glass bottle) using an ATI Axia80 force sensor for ground truth. The estimator trained on generated images achieves a measurement error within $0.1\,\text{N}$ of the model trained on real data, demonstrating robust generalization to complex materials (App.~\ref{app:push}).

\textbf{Real-time Pose Tracking.}\quad To evaluate dynamic performance, we track the pose of a printed cylinder, cross, and T-shape under continuous rotation and translation. Trained with our augmented data, the model tracks poses in real time at $10\,\text{Hz}$ (visualized in Fig.~\ref{fig:real_world}), highlighting its practical utility.

\textbf{Object Insertion.}\quad Using an XArm7 robot equipped with dual GelSight Mini sensors, we perform a high-precision insertion task with a tight $3\,\text{mm}$ tolerance. Across 20 trials per object, our strategy achieves success rates of \textbf{$90\%$} for the cylinder, \textbf{$85\%$} for the cross and T-shape, and \textbf{$75\%$} for a daily-object Type-C connector. These results demonstrate that models trained via single-image augmentation can master millimeter-level real-world manipulation (App.~\ref{app:insertion})

\begin{figure}[h]
    \centering
    \begin{minipage}[c]{0.48\textwidth}
\textbf{Imitation Learning.}\quad We conduct real-world imitation learning in a \textbf{Pick and Peg-in-Hole} Task to test the effectiveness of augmenting tactile data in imitation learning. This task requires the robot to reach and grasp a cylinder and insert it  into a target receptacle. As shown in Fig.~\ref{fig:real_world}, this task requires spatial generalizability across different locations, stable grasping during transport, and millimeter contact alignment during insertion. Since real-time contact masks are unavailable, we use only force-controlled tactile generation for data augmentation. 
    \end{minipage}
    \hfill
    \begin{minipage}[c]{0.5\textwidth}
        \centering
    \includegraphics[width=\textwidth]{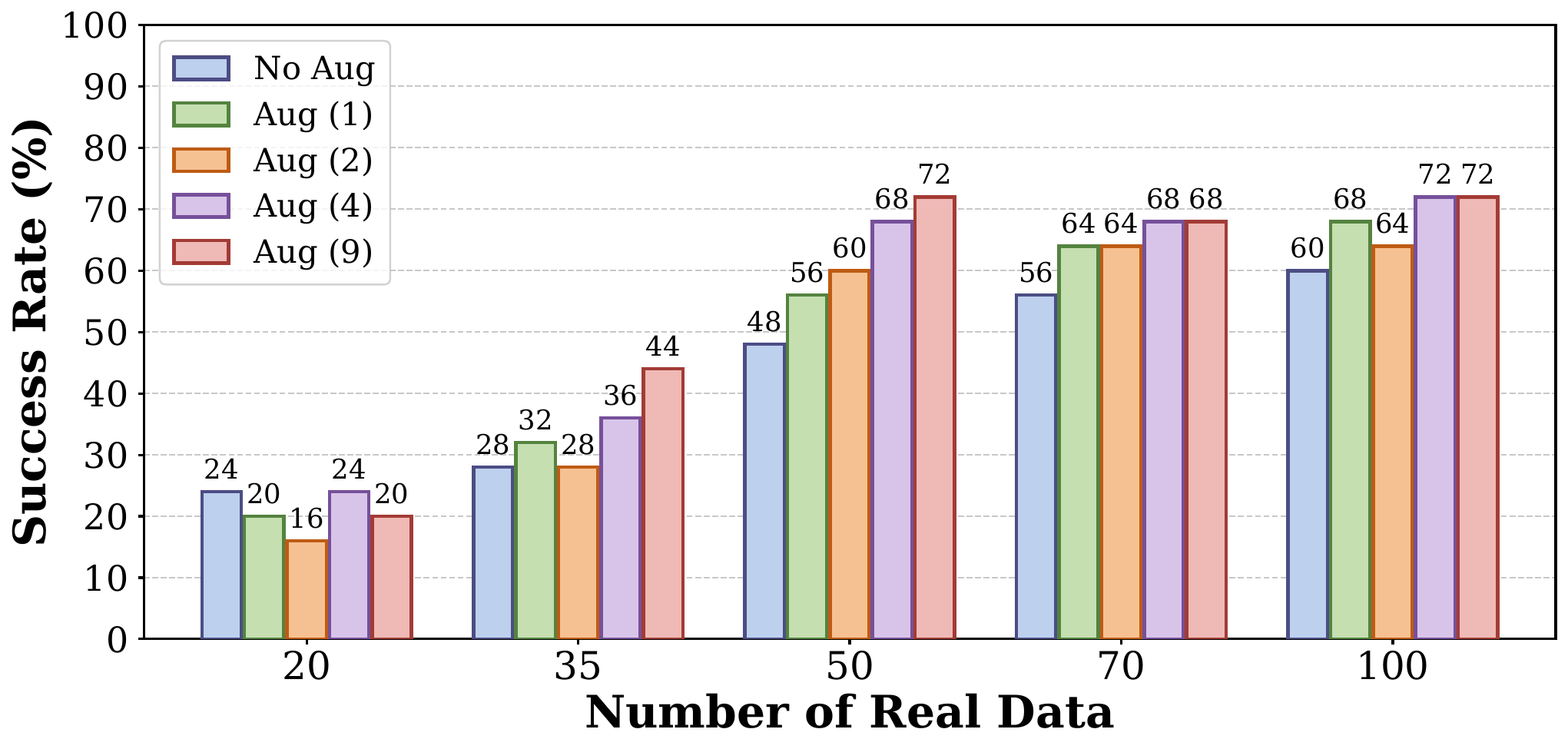}
    \caption{\textbf{Imitation Learning Results.} Success rates in the Pick and Peg-in-Hole task across different scales of real and augmented data.}
    \label{fig:reward}
    \end{minipage}
\end{figure}
To keep trajectories temporally consistent, we perform trajectory-level augmentation by applying a force offset sampled from $[-3,10]$ to each trajectory, with additional per-sample domain randomization from $\{-1,-0.5,0,0.5,1\}$. We train ACT-based~\cite{act} visuo-tactile policies with third-view RGB, wrist RGB, and tactile inputs using 20--100 real demonstrations and 1--9 augmented copies per trajectory, with details in App.~\ref{app:imitation}. Each policy is evaluated over 25 real-robot trials. Results show that augmentation is less effective with very limited data, where failures mainly come from poor spatial generalization. Once real data covers more object and hole locations, tactile augmentation improves robustness and success; with 70--100 demonstrations, performance saturates but augmentation still outperforms real-only training.

\vspace{-0.3cm}
\section{Conclusion and Limitation}\label{sec:conclusion}
\vspace{-0.1cm}
We presented \name, a two-stage conditional tactile generation framework that produces realistic and diverse tactile images from a single reference image by explicitly conditioning on contact forces and poses. Across downstream applications and real-world experiments, \name enables effective data augmentation, significantly outperforming prior approaches by capturing a wider range of dynamic variations. 

\textbf{Limitations.}\quad Despite these strengths, several limitations remain. First, the training pipeline requires manual alignment of contact masks, restricting scalability to diverse real-world tasks. Second, \name is tailored exclusively for vision-based tactile sensors, limiting generalizability to modalities like resistive or capacitive arrays. Lastly, the framework lacks robot action signals. Future work will focus on automating mask generation and coupling \name with robot interaction data to support richer conditioning for contact-rich manipulation.

\bibliography{main}

@InProceedings{obj_folder_bench,
    author    = {Gao, Ruohan and Dou, Yiming and Li, Hao and Agarwal, Tanmay and Bohg, Jeannette and Li, Yunzhu and Fei-Fei, Li and Wu, Jiajun},
    title     = {The ObjectFolder Benchmark: Multisensory Learning With Neural and Real Objects},
    booktitle = {CVPR},
    year      = {2023},
}

@inproceedings{controllable,
  title={Controllable visual-tactile synthesis},
  author={Gao, Ruihan and Yuan, Wenzhen and Zhu, Jun-Yan},
  booktitle={CVPR},
  year={2023}
}

@INPROCEEDINGS{texture,
  author={Li, Rui and Adelson, Edward H.},
  booktitle={CVPR}, 
  title={Sensing and Recognizing Surface Textures Using a GelSight Sensor}, 
  year={2013},}

@INPROCEEDINGS{normalflow,
    author={Huang, Hung-Jui and Kaess, Michael and Yuan, Wenzhen},
    booktitle={RAL}, 
    title={NormalFlow: Fast, Robust, and Accurate Contact-based Object 6DoF Pose Tracking with Vision-based Tactile Sensors}, 
    year={2024},
}

@INPROCEEDINGS{neuralfeels,
title={{N}eural feels with neural fields: {V}isuo-tactile perception for in-hand manipulation},
author={Suresh, Sudharshan and Qi, Haozhi and Wu, Tingfan and Fan, Taosha and Pineda, Luis and Lambeta, Mike and Malik, Jitendra and Kalakrishnan, Mrinal and Calandra, Roberto and Kaess, Michael and Ortiz, Joseph and Mukadam, Mustafa},
booktitle={Science Robotics},
year={2024},
}

@INPROCEEDINGS{touchgs,
  author={Swann, Aiden and Strong, Matthew and Do, Won Kyung and Camps, Gadiel Sznaier and Schwager, Mac and Kennedy, Monroe},
  booktitle={IROS}, 
  title={Touch-GS: Visual-Tactile Supervised 3D Gaussian Splatting}, 
  year={2024},}

@inproceedings{

mimictouch,

title={MimicTouch: Leveraging Multi-modal Human Tactile Demonstrations for Contact-rich Manipulation},

author={Kelin Yu and Yunhai Han and Qixian Wang and Vaibhav Saxena and Danfei Xu and Ye Zhao},

booktitle={CoRL},

year={2024},


}

@inproceedings{seehearfeel,
    title={See, Hear, and Feel: Smart Sensory Fusion for Robotic Manipulation},
    author={Hao Li and Yizhi Zhang and Junzhe Zhu and Shaoxiong Wang and Michelle A. Lee and Huazhe Xu and Edward Adelson and Li Fei-Fei and Ruohan Gao and Jiajun Wu},
    booktitle={CoRL},
    year={2022}
}

@inproceedings{tactilerl,
  title={Tactile-rl for insertion: Generalization to objects of unknown geometry},
  author={Dong, Siyuan and Jha, Devesh K and Romeres, Diego and Kim, Sangwoon and Nikovski, Daniel and Rodriguez, Alberto},
  booktitle={ICRA},
  year={2021}
}

@inproceedings{taxim,
  title={Taxim: An example-based simulation model for gelsight tactile sensors},
  author={Si, Zilin and Yuan, Wenzhen},
  booktitle={RAL},
  year={2022}
}

@inproceedings{tacto,
  author={Wang, Shaoxiong and Lambeta, Mike and Chou, Po-Wei and Calandra, Roberto},
  booktitle={RAL}, 
  title={TACTO: A Fast, Flexible, and Open-Source Simulator for High-Resolution Vision-Based Tactile Sensors}, 
  year={2022}}

@inproceedings{
difftactile,
title={{DIFFTACTILE}: A Physics-based Differentiable Tactile Simulator for Contact-rich Robotic Manipulation},
author={Zilin Si and Gu Zhang and Qingwei Ben and Branden Romero and Zhou Xian and Chao Liu and Chuang Gan},
booktitle={ICLR},
year={2024},
}

@article{braille,
  title={Learning to read braille: Bridging the tactile reality gap with diffusion models},
  author={Higuera, Carolina and Boots, Byron and Mukadam, Mustafa},
  journal={arXiv preprint arXiv:2304.01182},
  year={2023}
}

@article{sim2real,
  title={Reducing Tactile Sim2Real Domain Gaps via Deep Texture Generation Networks},
  author={Tudor Jianu, Daniel Fernandes Gomes, Shan Luo},
  journal={arXiv preprint arXiv:2112.01807},
  year={2021}
}

@inproceedings{
touchinganerf,
title={Touching a Ne{RF}: Leveraging Neural Radiance Fields for Tactile Sensory Data Generation},
author={Shaohong Zhong and Alessandro Albini and Oiwi Parker Jones and Perla Maiolino and Ingmar Posner},
booktitle={CoRL},
year={2022}
}

@inproceedings{tarf,
  title={Tactile-augmented radiance fields},
  author={Dou, Yiming and Yang, Fengyu and Liu, Yi and Loquercio, Antonio and Owens, Andrew},
  booktitle={CVPR},
  year={2024}
}

@inproceedings{
touchandgo,
title={Touch and Go: Learning from Human-Collected Vision and Touch},
author={Fengyu Yang and Chenyang Ma and Jiacheng Zhang and Jing Zhu and Wenzhen Yuan and Andrew Owens},
booktitle={NIPS},
year={2022},
}

@inproceedings{controlnet,
  title={Adding conditional control to text-to-image diffusion models},
  author={Zhang, Lvmin and Rao, Anyi and Agrawala, Maneesh},
  booktitle={CVPR},
  year={2023}
}

@article{reskin,
  title={Reskin: versatile, replaceable, lasting tactile skins},
  author={Bhirangi, Raunaq and Hellebrekers, Tess and Majidi, Carmel and Gupta, Abhinav},
  journal={arXiv preprint arXiv:2111.00071},
  year={2021}
}

@inproceedings{anyskin,
  title={Anyskin: Plug-and-play skin sensing for robotic touch},
  author={Bhirangi, Raunaq and Pattabiraman, Venkatesh and Erciyes, Enes and Cao, Yifeng and Hellebrekers, Tess and Pinto, Lerrel},
  booktitle={ICRA},
  year={2025},
}

@inproceedings{glove,
  title={Learning the signatures of the human grasp using a scalable tactile glove},
  author={Sundaram, Subramanian and Kellnhofer, Petr and Li, Yunzhu and Zhu, Jun-Yan and Torralba, Antonio and Matusik, Wojciech},
  booktitle={Nature},
  year={2019}
}

@misc{act,
      title={Learning Fine-Grained Bimanual Manipulation with Low-Cost Hardware}, 
      author={Tony Z. Zhao and Vikash Kumar and Sergey Levine and Chelsea Finn},
      year={2023},
      eprint={2304.13705},
      archivePrefix={arXiv},
      primaryClass={cs.RO},
      url={https://arxiv.org/abs/2304.13705}, 
}

@inproceedings{diffusionpolicy,
  title={Diffusion Policy: Visuomotor Policy Learning via Action Diffusion},
  author={Chi, Cheng and Feng, Siyuan and Du, Yilun and Xu, Zhenjia and Cousineau, Eric and Burchfiel, Benjamin and Song, Shuran},
  booktitle={Robotics: Science and Systems},
  year={2023}
}

@inproceedings{dp3,
  title={{3D Diffusion Policy}: Generalizable Visuomotor Policy Learning via Simple 3D Representations},
  author={Ze, Yanjie and Zhang, Gu and Zhang, Kangning and Hu, Chenyuan and Wang, Muhan and Xu, Huazhe},
  booktitle={Robotics: Science and Systems},
  year={2024}
}

@inproceedings{3dvitac,
title={3D ViTac:Learning Fine-Grained Manipulation with Visuo-Tactile Sensing},
author={Huang, Binghao and Wang, Yixuan and Yang, Xinyi and Luo, Yiyue and Li, Yunzhu},
booktitle={CoRL},
year={2024}
}

@inproceedings{gelsight,
  title={Gelsight: High-resolution robot tactile sensors for estimating geometry and force},
  author={Yuan, Wenzhen and Dong, Siyuan and Adelson, Edward H},
  booktitle={Sensors},
  year={2017},
}

@inproceedings{gelslim,
  title={Gelslim 3.0: High-resolution measurement of shape, force and slip in a compact tactile-sensing finger},
  author={Taylor, Ian H and Dong, Siyuan and Rodriguez, Alberto},
  booktitle={ICRA},
  year={2022}}

@inproceedings{digit,
  title={Digit: A novel design for a low-cost compact high-resolution tactile sensor with application to in-hand manipulation},
  author={Lambeta, Mike and Chou, Po-Wei and Tian, Stephen and Yang, Brian and Maloon, Benjamin and Most, Victoria Rose and Stroud, Dave and Santos, Raymond and Byagowi, Ahmad and Kammerer, Gregg and others},
  booktitle={RAL},
  year={2020}}

@inproceedings{9dtact,
  title={9dtact: A compact vision-based tactile sensor for accurate 3d shape reconstruction and generalizable 6d force estimation},
  author={Lin, Changyi and Zhang, Han and Xu, Jikai and Wu, Lei and Xu, Huazhe},
  booktitle={RAL},
  year={2023}
}

@article{ifem,
  title={Dense Tactile Force Distribution Estimation using GelSlim and inverse FEM. arXiv 2018},
  author={Ma, D and Donlon, E and Dong, S and Rodriguez, A},
  journal={arXiv preprint arXiv:1810.04621},
  year={2018}
}

@inproceedings{feelanyforce,
  title={FeelAnyForce: Estimating contact force feedback from tactile sensation for vision-based tactile sensors},
  author={Shahidzadeh, Amir-Hossein and Caddeo, Gabriele M and Alapati, Koushik and Natale, Lorenzo and Ferm{\"u}ler, Cornelia and Aloimonos, Yiannis},
  booktitle={ICRA},
  year={2025}}

@inproceedings{dreamfusion,
  title={Tactile DreamFusion: Exploiting tactile sensing for 3D generation},
  author={Gao, Ruihan and Deng, Kangle and Yang, Gengshan and Yuan, Wenzhen and Zhu, Jun-Yan},
  booktitle={NIPS},
  year={2024}
}

@inproceedings{hardness,
   title={Shape-independent hardness estimation using deep learning and a GelSight tactile sensor},
   booktitle={ICRA},
   author={Yuan, Wenzhen and Zhu, Chenzhuo and Owens, Andrew and Srinivasan, Mandayam A. and Adelson, Edward H.},
   year={2017}}

@article{liquid,
  title={Understanding dynamic tactile sensing for liquid property estimation},
  author={Huang, Hung-Jui and Guo, Xiaofeng and Yuan, Wenzhen},
  journal={arXiv preprint arXiv:2205.08771},
  year={2022}
}

@inproceedings{slip,
  author={Li, Jianhua and Dong, Siyuan and Adelson, Edward},
  booktitle={ICRA}, 
  title={Slip Detection with Combined Tactile and Visual Information}, 
  year={2018}}

@inproceedings{feeling,
  title={More than a feeling: Learning to grasp and regrasp using vision and touch},
  author={Calandra, Roberto and Owens, Andrew and Jayaraman, Dinesh and Lin, Justin and Yuan, Wenzhen and Malik, Jitendra and Adelson, Edward H and Levine, Sergey},
  booktitle={RAL},
  year={2018}}

@article{grasping,
  title={The feeling of success: Does touch sensing help predict grasp outcomes?},
  author={Calandra, Roberto and Owens, Andrew and Upadhyaya, Manu and Yuan, Wenzhen and Lin, Justin and Adelson, Edward H and Levine, Sergey},
  journal={arXiv preprint arXiv:1710.05512},
  year={2017}
}

@inproceedings{deformable,
  title={Learning generalizable vision-tactile robotic grasping strategy for deformable objects via transformer},
  author={Han, Yunhai and Yu, Kelin and Batra, Rahul and Boyd, Nathan and Mehta, Chaitanya and Zhao, Tuo and She, Yu and Hutchinson, Seth and Zhao, Ye},
  booktitle={TMECH},
  year={2024}}

@inproceedings{
rotation,
title={General In-hand Object Rotation with Vision and Touch},
author={Haozhi Qi and Brent Yi and Sudharshan Suresh and Mike Lambeta and Yi Ma and Roberto Calandra and Jitendra Malik},
booktitle={CoRL},
year={2023}
}

@inproceedings{packing,
  title={Tactile-based insertion for dense box-packing},
  author={Dong, Siyuan and Rodriguez, Alberto},
  booktitle={IROS},
  year={2019}}

@article{robopack,
  title={Robopack: Learning tactile-informed dynamics models for dense packing},
  author={Ai, Bo and Tian, Stephen and Shi, Haochen and Wang, Yixuan and Tan, Cheston and Li, Yunzhu and Wu, Jiajun},
  journal={arXiv preprint arXiv:2407.01418},
  year={2024}
}

@article{t3,
  title={Transferable tactile transformers for representation learning across diverse sensors and tasks},
  author={Zhao, Jialiang and Ma, Yuxiang and Wang, Lirui and Adelson, Edward H},
  journal={arXiv preprint arXiv:2406.13640},
  year={2024}
}

@inproceedings{
sensorinv,
title={Sensor-Invariant Tactile Representation},
author={Harsh Gupta and Yuchen Mo and Shengmiao Jin and Wenzhen Yuan},
booktitle={ICLR},
year={2025},
}

@inproceedings{
    sparsh,
    title={Sparsh: Self-supervised touch representations for vision-based tactile sensing},
    author={Carolina Higuera and Akash Sharma and Chaithanya Krishna Bodduluri and Taosha Fan and Patrick Lancaster and Mrinal Kalakrishnan and Michael Kaess and Byron Boots and Mike Lambeta and Tingfan Wu and Mustafa Mukadam},
    booktitle={CoRL},
    year={2024}
}

@inproceedings{objectforlder,
  title = {ObjectFolder: A Dataset of Objects with Implicit Visual, Auditory, and Tactile Representations},
  author = {Gao, Ruohan and Chang, Yen-Yu and Mall, Shivani and Fei-Fei, Li and Wu, Jiajun},
  booktitle = {CoRL},
  year = {2021}
}

@InProceedings{visgel,
	author = {Li, Yunzhu and Zhu, Jun-Yan and Tedrake, Russ and Torralba, Antonio},
	title = {Connecting Touch and Vision via Cross-Modal Prediction},
	booktitle = {CVPR},
	year = {2019}
}

@inproceedings{objectfolder2,
  title = {ObjectFolder 2.0: A Multisensory Object Dataset for Sim2Real Transfer},
  author = {Gao, Ruohan and Si, Zilin and Chang, Yen-Yu and Clarke, Samuel and Bohg, Jeannette and Fei-Fei, Li and Yuan, Wenzhen and Wu, Jiajun},
  booktitle = {CVPR},
  year = {2022},
}

@inproceedings{
anytouch,
title={AnyTouch: Learning Unified Static-Dynamic Representation across Multiple Visuo-tactile Sensors},
author={Ruoxuan Feng and Jiangyu Hu and Wenke Xia and TianciGao and Ao Shen and Yuhao Sun and Bin Fang and Di Hu},
booktitle={ICLR},
year={2025}
}

@inproceedings{pbr,
  title={Simulation of vision-based tactile sensors using physics based rendering},
  author={Agarwal, Arpit and Man, Timothy and Yuan, Wenzhen},
  booktitle={ICRA},
  year={2021},
}

@article{sana,
  title={Sana: Efficient high-resolution image synthesis with linear diffusion transformers},
  author={Xie, Enze and Chen, Junsong and Chen, Junyu and Cai, Han and Tang, Haotian and Lin, Yujun and Zhang, Zhekai and Li, Muyang and Zhu, Ligeng and Lu, Yao and others},
  journal={arXiv preprint arXiv:2410.10629},
  year={2024}
}

@inproceedings{dit,
  title={Scalable diffusion models with transformers},
  author={Peebles, William and Xie, Saining},
  booktitle={CVPR},
  year={2023}
}

@article{ddim,
  title={Denoising diffusion implicit models},
  author={Song, Jiaming and Meng, Chenlin and Ermon, Stefano},
  journal={arXiv preprint arXiv:2010.02502},
  year={2020}
}

@inproceedings{binding,
  title={Binding touch to everything: Learning unified multimodal tactile representations},
  author={Yang, Fengyu and Feng, Chao and Chen, Ziyang and Park, Hyoungseob and Wang, Daniel and Dou, Yiming and Zeng, Ziyao and Chen, Xien and Gangopadhyay, Rit and Owens, Andrew and others},
  booktitle={CVPR},
  year={2024}
}

@article{touch2touch,
  title={Touch2touch: Cross-modal tactile generation for object manipulation},
  author={Rodriguez, Samanta and Dou, Yiming and Oller, Miquel and Owens, Andrew and Fazeli, Nima},
  journal={arXiv preprint arXiv:2409.08269},
  year={2024}
}

@article{pixart,
  title={Pixart-$\{$$\backslash$delta$\}$: Fast and controllable image generation with latent consistency models},
  author={Chen, Junsong and Wu, Yue and Luo, Simian and Xie, Enze and Paul, Sayak and Luo, Ping and Zhao, Hang and Li, Zhenguo},
  journal={arXiv preprint arXiv:2401.05252},
  year={2024}
}

@article{vit,
  title={An image is worth 16x16 words: Transformers for image recognition at scale},
  author={Dosovitskiy, Alexey and Beyer, Lucas and Kolesnikov, Alexander and Weissenborn, Dirk and Zhai, Xiaohua and Unterthiner, Thomas and Dehghani, Mostafa and Minderer, Matthias and Heigold, Georg and Gelly, Sylvain and others},
  journal={arXiv preprint arXiv:2010.11929},
  year={2020}
}

@article{dinov2,
  title={Dinov2: Learning robust visual features without supervision},
  author={Oquab, Maxime and Darcet, Timoth{\'e}e and Moutakanni, Th{\'e}o and Vo, Huy and Szafraniec, Marc and Khalidov, Vasil and Fernandez, Pierre and Haziza, Daniel and Massa, Francisco and El-Nouby, Alaaeldin and others},
  journal={arXiv preprint arXiv:2304.07193},
  year={2023}
}

@inproceedings{imagenet,
  title={Imagenet: A large-scale hierarchical image database},
  author={Deng, Jia and Dong, Wei and Socher, Richard and Li, Li-Jia and Li, Kai and Fei-Fei, Li},
  booktitle={CVPR},
  year={2009}}

@article{CGAN,
  title={Conditional generative adversarial nets},
  author={Mirza, Mehdi and Osindero, Simon},
  journal={arXiv preprint arXiv:1411.1784},
  year={2014}
}

@inproceedings{CGAN1,
  title={Generative adversarial text to image synthesis},
  author={Reed, Scott and Akata, Zeynep and Yan, Xinchen and Logeswaran, Lajanugen and Schiele, Bernt and Lee, Honglak},
  booktitle={ICML},
  year={2016}
}

@inproceedings{CGAN2,
  title={Image-to-image translation with conditional adversarial networks},
  author={Isola, Phillip and Zhu, Jun-Yan and Zhou, Tinghui and Efros, Alexei A},
  booktitle={CVPR},
  year={2017}
}

@inproceedings{CGAN3,
  title={Unpaired image-to-image translation using cycle-consistent adversarial networks},
  author={Zhu, Jun-Yan and Park, Taesung and Isola, Phillip and Efros, Alexei A},
  booktitle={CVPR},
  year={2017}
}

@inproceedings{CVAE,
  title={Learning structured output representation using deep conditional generative models},
  author={Sohn, Kihyuk and Lee, Honglak and Yan, Xinchen},
  booktitle={NIPS},
  year={2015}
}

@inproceedings{CVAE1,
  title={CVAE-GAN: fine-grained image generation through asymmetric training},
  author={Bao, Jianmin and Chen, Dong and Wen, Fang and Li, Houqiang and Hua, Gang},
  booktitle={CVPR},
  year={2017}
}

@inproceedings{CVAE2,
  title={Disentangled representation learning gan for pose-invariant face recognition},
  author={Tran, Luan and Yin, Xi and Liu, Xiaoming},
  booktitle={CVPR},
  year={2017}
}

@inproceedings{CVAE3,
  title={Attribute2image: Conditional image generation from visual attributes},
  author={Yan, Xinchen and Yang, Jimei and Sohn, Kihyuk and Lee, Honglak},
  booktitle={ECCV},
  year={2016},
}

@article{ddpm,
  title={Denoising diffusion probabilistic models},
  author={Ho, Jonathan and Jain, Ajay and Abbeel, Pieter},
  journal={NIPS},
  year={2020}
}

@article{score,
  title={Score-based generative modeling through stochastic differential equations},
  author={Song, Yang and Sohl-Dickstein, Jascha and Kingma, Diederik P and Kumar, Abhishek and Ermon, Stefano and Poole, Ben},
  journal={arXiv preprint arXiv:2011.13456},
  year={2020}
}

@inproceedings{stablediffusion,
  title={High-resolution image synthesis with latent diffusion models},
  author={Rombach, Robin and Blattmann, Andreas and Lorenz, Dominik and Esser, Patrick and Ommer, Bj{\"o}rn},
  booktitle={CVPR},
  year={2022}
}

@article{GanL1,
  title={Progressive growing of gans for improved quality, stability, and variation},
  author={Karras, Tero and Aila, Timo and Laine, Samuli and Lehtinen, Jaakko},
  journal={arXiv preprint arXiv:1710.10196},
  year={2017}
}

@article{GanL2,
  title={Understanding the limitations of conditional generative models},
  author={Fetaya, Ethan and Jacobsen, J{\"o}rn-Henrik and Grathwohl, Will and Zemel, Richard},
  journal={arXiv preprint arXiv:1906.01171},
  year={2019}
}

@inproceedings{GanL3,
  title={A survey on generative adversarial networks: Variants, applications, and training},
  author={Jabbar, Abdul and Li, Xi and Omar, Bourahla},
  booktitle={ACM CSUR},
  year={2021},
}

@inproceedings{GanL4,
  title={Deep generative models: Survey},
  author={Oussidi, Achraf and Elhassouny, Azeddine},
  booktitle={ISCV},
  year={2018},
}

@inproceedings{texttoucher,
  title={Texttoucher: Fine-grained text-to-touch generation},
  author={Tu, Jiahang and Fu, Hao and Yang, Fengyu and Zhao, Hanbin and Zhang, Chao and Qian, Hui},
  booktitle={AAAI},
  year={2025}
}

@inproceedings{generating,
  title={Generating Visual Scenes from Touch},
  author={Yang, Fengyu and Zhang, Jiacheng and Owens, Andrew},
  booktitle={ICCV},
  year={2023},
}

@inproceedings{data_aug1,
  title={The impact of data augmentation on tactile-based object classification using deep learning approach},
  author={Maus, Philip and Kim, Jaeseok and Nocentini, Olivia and Bashir, Muhammad Zain and Cavallo, Filippo},
  booktitle={Sensors},
  year={2022}
}

@inproceedings{data_aug2,
  title={Geometric Transformation: Tactile Data Augmentation for Robotic Learning},
  author={Yan, Gang and Yuyeol, Jun and Funabashi, Satoshi and Tomo, Tito Pradhono and Somlor, Sophon and Schmitz, Alexander and Sugano, Shigeki},
  booktitle={ICDL},
  year={2023}
}

@inproceedings{canny,
  title={A computational approach to edge detection},
  author={Canny, John},
  booktitle={TPAMI},
  year={1986},
}

@article{admw,
  title={Decoupled weight decay regularization},
  author={Loshchilov, Ilya and Hutter, Frank},
  journal={arXiv preprint arXiv:1711.05101},
  year={2017}
}

@inproceedings{10611667, author={Shahidzadeh, Amir-Hossein and Yoo, Seong Jong and Mantripragada, Pavan and Singh, Chahat Deep and Fermüller, Cornelia and Aloimonos, Yiannis}, booktitle={ICRA}, title={AcTExplore: Active Tactile Exploration on Unknown Objects}, year={2024}, }

@inproceedings{cable,
  title={Cable manipulation with a tactile-reactive gripper},
  author={She, Yu and Wang, Shaoxiong and Dong, Siyuan and Sunil, Neha and Rodriguez, Alberto and Adelson, Edward},
  booktitle={IJRR},
  year={2021}
}

@article{tactilefilter,
  title={Tactile-filter: Interactive tactile perception for part mating},
  author={Ota, Kei and Jha, Devesh K and Tung, Hsiao-Yu and Tenenbaum, Joshua B},
  journal={arXiv preprint arXiv:2303.06034},
  year={2023}
}

@inproceedings{tac2pose,
  title={Tac2pose: Tactile object pose estimation from the first touch},
  author={Bauza, Maria and Bronars, Antonia and Rodriguez, Alberto},
  booktitle={IJRR},
  year={2023}
}

@article{tacex,
  title={TacEx: GelSight Tactile Simulation in Isaac Sim--Combining Soft-Body and Visuotactile Simulators},
  author={Nguyen, Duc Huy and Schneider, Tim and Duret, Guillaume and Kshirsagar, Alap and Belousov, Boris and Peters, Jan},
  journal={arXiv preprint arXiv:2411.04776},
  year={2024}
}

@article{tactileoverview,
  title={Tactile robotics: An outlook},
  author={Luo, Shan and Lepora, Nathan F and Yuan, Wenzhen and Althoefer, Kaspar and Cheng, Gordon and Dahiya, Ravinder},
  journal={T-RO},
  year={2025},
}

@article{anytouch2,
  title={AnyTouch 2: General Optical Tactile Representation Learning For Dynamic Tactile Perception},
  author={Feng, Ruoxuan and Zhou, Yuxuan and Mei, Siyu and Zhou, Dongzhan and Wang, Pengwei and Cui, Shaowei and Fang, Bin and Yao, Guocai and Hu, Di},
  journal={arXiv preprint arXiv:2602.09617},
  year={2026}
}

@article{taccel,
  title={Taccel: Scaling up vision-based tactile robotics via high-performance gpu simulation},
  author={Li, Yuyang and Du, Wenxin and Yu, Chang and Li, Puhao and Zhao, Zihang and Liu, Tengyu and Jiang, Chenfanfu and Zhu, Yixin and Huang, Siyuan},
  journal={arXiv preprint arXiv:2504.12908},
  year={2025}
}

@article{aloha,
  title={Aloha 2: An enhanced low-cost hardware for bimanual teleoperation},
  author={Aldaco, Jorge and Armstrong, Travis and Baruch, Robert and Bingham, Jeff and Chan, Sanky and Draper, Kenneth and Dwibedi, Debidatta and Finn, Chelsea and Florence, Pete and Goodrich, Spencer and others},
  journal={arXiv preprint arXiv:2405.02292},
  year={2024}
}

\appendix
\clearpage

\section*{Appendix Overview}

\noindent \textbf{Details of Vision-Based Tactile Sensors} \dotfill \textbf{Section~\ref{app:tactile_details}} \\
\noindent \textbf{Implementation Details} \dotfill \textbf{Section~\ref{app:implementation}} \\
\hspace*{0.75em} Generative Model Setup \dotfill Sec.~\ref{app:model_setup} \\
\hspace*{0.75em} Training Configuration \dotfill Sec.~\ref{app:train} \\
\hspace*{0.75em} Inference Configuration \dotfill Sec.~\ref{app:inference} \\
\hspace*{0.75em} Downstream Tasks Setup \dotfill Sec.~\ref{app:downstream_setup} \\
\hspace*{0.75em} Metrics \dotfill Sec.~\ref{app:metrics} \\
\hspace*{0.75em} Evaluation Dataset Composition \dotfill Sec.~\ref{app:dataset_composition} \\[0.5em]
\noindent \textbf{Details of Baselines} \dotfill \textbf{Section~\ref{app:baselines}} \\
\hspace*{0.75em} Hybrid Force-Pose Conditional Diffusion Model \dotfill Sec.~\ref{app:hybrid_baseline} \\
\hspace*{0.75em} Separate Force-Pose Conditional Diffusion Model \dotfill Sec.~\ref{app:separate_baseline} \\[0.5em]
\noindent \textbf{Network Architectures for Downstream Tasks} \dotfill \textbf{Section~\ref{app:downstream_architectures}} \\
\hspace*{0.75em} Object Classifier Architecture \dotfill Sec.~\ref{app:classifier} \\
\hspace*{0.75em} Force Estimator Architecture \dotfill Sec.~\ref{app:force_estimator} \\
\hspace*{0.75em} Pose Estimator Architecture \dotfill Sec.~\ref{app:pose_estimator} \\[0.5em]
\noindent \textbf{Additional Experiments} \dotfill \textbf{Section~\ref{app:addition_exp}} \\
\hspace*{0.75em} Physical Consistency and Coverage Validation \dotfill Sec.~\ref{app:physical_validation} \\
\hspace*{0.75em} Manipulation Evaluation Details \dotfill Sec.~\ref{app:manipulation_details} \\
\hspace*{0.75em} Single-Reference Inference and Practical Costs \dotfill Sec.~\ref{app:single_reference} \\
\hspace*{0.75em} Comparison with Vision to Tactile Generation \dotfill Sec.~\ref{app:vis2tac} \\
\hspace*{0.75em} Robustness Validation of Contact Mask Alignment \dotfill Sec.~\ref{app:mask} \\
\hspace*{0.75em} Details of Object Classification in \textsc{Text2Tac} \dotfill Sec.~\ref{app:text2tac_class} \\
\hspace*{0.75em} Impact of Data Composition on Force Estimation \dotfill Sec.~\ref{app:impact} \\
\hspace*{0.75em} Force Estimation with Force-Control Data Augmentation \dotfill Sec.~\ref{app:force} \\
\hspace*{0.75em} Complete Comparison of Pose Estimation Performance \dotfill Sec.~\ref{app:full_comparison} \\
\hspace*{0.75em} Details of Object Weighting \dotfill Sec.~\ref{app:push} \\
\hspace*{0.75em} Details of Object Insertion \dotfill Sec.~\ref{app:insertion} \\
\hspace*{0.75em} Details of Imitation Learning \dotfill Sec.~\ref{app:imitation} \\[0.5em]
\noindent \textbf{Failure Analysis} \dotfill \textbf{Section~\ref{app:failure}} \\[0.5em]
\noindent \textbf{Addition Visualizations} \dotfill \textbf{Section~\ref{app:vis}} \\
\hspace*{0.75em} Complete Visualization Results \dotfill Sec.~\ref{app:full_vis} \\
\hspace*{0.75em} Visualization of Generated Image using Force-Control Generation \dotfill Sec.~\ref{app:force_only} \\
\hspace*{0.75em} Visualization of Depth Map \dotfill Sec.~\ref{app:depth} \\
\hspace*{0.75em} Object and Tactile Image Visualization for Classification \dotfill Sec.~\ref{app:class} \\
\hspace*{0.75em} Sensitivity of Tactile Images to Force Variations \dotfill Sec.~\ref{app:force_sensitivity} \\
\hspace*{0.75em} Visualization of different sensor samples \dotfill Sec.~\ref{app:sensor} \\
\hspace*{0.75em} Visualization of a special case requiring generation \dotfill Sec.~\ref{app:special} \\
\hspace*{0.75em} Visualization Results in the Simulator \dotfill Sec.~\ref{app:vis_taxim} \\
\hspace*{0.75em} Visualization of the Type-C Connector Insertion Task \dotfill Sec.~\ref{app:usb}

\newpage

\section{Details of Vision-Based Tactile Sensors}
\label{app:tactile_details}

To provide a comprehensive background on the hardware modalities, this section details the underlying mechanisms and applications of modern tactile sensing technologies. Various tactile sensors have been systematically designed for diverse robotic scenarios, primary categories including magnetic~\cite{reskin, anyskin}, piezo-resistive~\cite{glove, 3dvitac}, and vision-based~\cite{gelsight, gelslim, digit, 9dtact} sensors. Among these distinct modalities, vision-based tactile sensors have gained significant momentum within the community. By utilizing internal miniature cameras to capture high-resolution RGB images that record elastomeric gel deformations, these sensors deliver exceptional spatial resolution and rich surface appearance cues that surpass conventional tactile alternatives.

Specifically, vision-based tactile sensors consist of a soft elastomer gel, a textured background layer, an LED illumination module, and an onboard camera to capture gel deformation. The background layer may include markers to improve force estimation~\cite{gelslim}, while illumination affects the rendered shading and contact patterns. Variations in gel stiffness, fabrication, and lighting can therefore lead to substantial appearance differences even between identical sensors~\cite{t3, sensorinv}. Furthermore, contact force estimation from tactile sensors via soft elastomer gel deformation is highly non-trivial. The mapping requires complex finite-element modeling due to varying material properties such as stiffness, thickness, and damping~\cite{ifem}. The 1-D normal and 2-D shear components of the 3D contact force vector do not map directly to the measured force distribution, as they are entangled with local geometry, contact patch shape, and pose, meaning identical ground-truth forces can produce entirely different deformation patterns.

Consequently, vision-based sensors have become instrumental across a wide spectrum of robotic capabilities. In perception-driven tasks, they are extensively applied to material property classification~\cite{liquid, hardness}, high-fidelity 3D reconstruction~\cite{normalflow, neuralfeels, touchgs, dreamfusion, 10611667}, real-time slip detection~\cite{slip}, and object pose estimation~\cite{normalflow}. Meanwhile, in contact-rich manipulation tasks, they provide critical feedback for robust grasping~\cite{feeling, grasping, deformable}, precise insertion~\cite{mimictouch, seehearfeel, tactilerl}, liquid pouring~\cite{seehearfeel}, in-hand object rotation~\cite{rotation}, and dense packing~\cite{mimictouch, packing, robopack}.

\section{Implementation Details}\label{app:implementation}

\subsection{Generative Model Setup}\label{app:model_setup}
For force-control generation (Stage 1), the latent autoencoder is instantiated using SANA~\cite{sana}, and the diffusion model is realized via a Diffusion Transformer (DiT~\cite{dit}) with DDIM sampling~\cite{ddim}. We train this force-control generator component using 20,000 tactile images with corresponding 3D force vectors from FeelAnyForce~\cite{feelanyforce}. For pose-control generation (Stage 2), the pose-guided adapter is trained using 7,000 tactile images, where each object contributes 300 unique contact poses paired with 3D force vectors. For training the baseline models, the \textsc{Hybrid} model uses 7,000 samples, and the \textsc{Separate}-Control pipeline uses 20,000 samples for force control and 7,000 for pose control. 

\subsection{Training Configuration}\label{app:train}
Both the force-control generation component and the pose-control generation component are trained using the AdamW~\cite{admw} optimizer and a cosine annealing learning rate scheduler. For the force-control generator, the learning rate is annealed from an initial value of $1 \times 10^{-4}$ to a final value of $1 \times 10^{-5}$. For the pose-control generator, the learning rate decays from $1 \times 10^{-5}$ to $1 \times 10^{-6}$. Each model is trained for 75{,}000 steps on a single NVIDIA RTX A5000 GPU with a batch size of 4. The loss function used for training is a weighted combination of L1 loss and mean squared error (MSE): $0.5 \times \mathcal{L}_{\text{L1}} + 0.5 \times \mathcal{L}_{\text{MSE}}$.

Specifically, the first stage (force-control generation) is optimized using a network architecture consisting of 147.53M parameters. It is trained for 50 epochs, with each epoch requiring approximately 1,500 seconds, resulting in a training duration of roughly 20.83 hours. The second stage (pose-control generation) utilizes a network architecture comprising 222.93M parameters. This stage converges within 30 epochs, with each epoch taking approximately 550 seconds, totaling 4.58 hours. Consequently, the entire framework maintains a compact parameter footprint of 370.46M parameters in total, and the end-to-end training process is successfully completed within approximately 25.41 hours.

\subsection{Inference Configuration}\label{app:inference}
We perform inference on a single NVIDIA RTX A6000 GPU with a batch size of 128. ControlTac achieves a throughput of 6.5 tactile images per second, while the \textsc{Hybrid} and \textsc{Separate} baseline methods reach 7 and 3.7 tactile images per second, respectively. Since the VAE decoding process is the most memory-intensive step, all three methods exhibit a peak GPU memory usage of 29.97 GB.

\subsection{Downstream Tasks Setup}\label{app:downstream_setup}
To systematically evaluate the generated data, we set up three downstream tasks. For force estimation, we adopt the framework from FeelAnyForce~\cite{feelanyforce}, utilizing a ViT encoder~\cite{vit} pretrained on DINOv2~\cite{dinov2}. For pose estimation, we adapt this ViT/DINOv2 architecture to output a 3-DoF contact pose. Finally, for object classification across six unseen objects, we benchmark performance across a plain CNN, an unpretrained ViT~\cite{vit}, and an ImageNet-pretrained ViT~\cite{imagenet}.

\subsection{Metrics}~\label{app:metrics}
We evaluate our models using several commonly used metrics, including mean squared error (MSE), L1 loss, mean absolute error (MAE), and structural similarity index measure (SSIM). Specifically, the following metrics are reported:
\begin{itemize}
    \item \textbf{Mean Squared Error (MSE):} $$ \mathrm{MSE} = \frac{1}{n} \sum_{i=1}^n (y_i - \hat{y}_i)^2 $$
    
    \item \textbf{L1 Loss:} 
    $$ \mathrm{L1} = \frac{1}{n} \sum_{i=1}^n |y_i - \hat{y}_i| $$
    
    \item \textbf{Mean Absolute Error (MAE):} 
    $$ \mathrm{MAE} = \frac{1}{n} \sum_{i=1}^n |y_i - \hat{y}_i| $$
    
    \item \textbf{Structural Similarity Index Measure (SSIM):} 
    $$ \mathrm{SSIM}(x, y) = \frac{(2 \mu_x \mu_y + C_1)(2 \sigma_{xy} + C_2)}{(\mu_x^2 + \mu_y^2 + C_1)(\sigma_x^2 + \sigma_y^2 + C_2)} $$
    
\end{itemize}

\subsection{Evaluation Dataset Composition}
\label{app:dataset_composition}
We evaluate our method using SSIM and pixel-wise MSE on FeelAnyForce~\cite{feelanyforce} test data. To assess zero-shot generalization on highly diverse and challenging geometries, we further incorporate objects from the large-scale AnyTouch2~\cite{anytouch2} dataset. The resulting evaluation set comprises twelve objects: six seen objects (evaluated under novel contact poses and forces), two unseen objects from FeelAnyForce, and four unseen objects from AnyTouch2.

\section{Details of Baselines}\label{app:baselines}

\subsection{Hybrid Force-Pose Conditional Diffusion Model:}\label{app:hybrid_baseline}
In this approach, we train a diffusion model $ \mathbf{y} = \mathcal{D}(\mathcal{F}_h(\mathbf{z^{(x)}}, \mathbf{z^{(c)}}, \mathbf{\Delta f})) $. Here, the latent representation of initial tactile image $ \mathbf{z^{(x)}} $, contact mask $ \mathbf{z^{(c)}} $, and target force change $ \mathbf{\Delta f} $ are simultaneously input into the diffusion model $ \mathcal{F}_h(\cdot) $, which is then passed into the autodecoder $ \mathcal{D}(\cdot) $ to generate the output $\mathbf{y}$.

\subsection{Separate Force-Pose Conditional Diffusion Model:}\label{app:separate_baseline}  
In the first stage, we follow the previous force-control generator method by inputting the latent representation of the initial tactile image $\mathbf{z^{(x)}}$ and the target force change $\mathbf{\Delta f}$ into the force-control generator $\mathcal{F}_f(\cdot)$ to produce a latent representation of the tactile image $\mathbf{z^{(x')}}=\mathcal{F}_f(\mathbf{z^{(x)},\mathbf{\Delta f}})$ that satisfies the target force. In the second stage, this generated latent representation $\mathbf{z^{(x')}}$, along with the latent representation of the contact mask $\mathbf{z^{(c)}}$, is input into the pose-control generator $\mathcal{F}_p(\cdot)$. The output $\mathbf{z^{(y)}=\mathcal{F}}_p(\mathbf{z^{(x')},z^{(c)}})$, which satisfies both the target force and contact pose, is then decoded to produce the final tactile image $\mathbf{y}=\mathcal{D}(\mathbf{z^{(y)}})$.

\section{Network Architectures for Downstream Tasks}
\label{app:downstream_architectures}

\subsection{Object Classifier Architecture}
\label{app:classifier}
To evaluate object classification, we employ two architecture-agnostic models: a custom Convolutional Neural Network (CNN) and a Vision Transformer (ViT-B/16). Their architectural configurations are summarized in Table~\ref{tab:classifier-comparison}.

\begin{table}[H]
\centering
\caption{Architectural comparison of the classification backbones.}
\label{tab:classifier-comparison}
\small
\begin{tabular}{l l l}
\toprule
\textbf{Component} & \textbf{CNN (Custom)} & \textbf{ViT-B/16} \\
\midrule
\textbf{Backbone} & 5-layer Conv (256 ch) & 12-layer Trans. (768-d) \\
\textbf{Bottleneck} & Flatten (50,176) & CLS Token (768) \\
\textbf{Head} & MLP ($512 \to 6$) & Linear ($768 \to 6$) \\
\textbf{Pretraining} & None & ImageNet-21k~\cite{imagenet} / None \\
\bottomrule
\end{tabular}
\end{table}

\subsection{Force Estimator Architecture}
\label{app:force_estimator}
Following the mathematical framework of FeelAnyForce~\cite{feelanyforce}, the force estimator maps an input tactile image to a continuous 3D force vector $\hat{F} \in \mathbb{R}^3$ (normal and 2D shear forces). The backbone utilizes a large-scale Vision Transformer (ViT)~\cite{vit} pre-trained via the self-supervised DINOv2 paradigm~\cite{dinov2}. DINOv2 is chosen for its robustness against local texture noise, forcing the model to rely on structural macro-deformations. The encoded latent features are then passed through a dense Multi-Layer Perceptron (MLP) head to regress $\hat{F}$.

\subsection{Pose Estimator Architecture}
\label{app:pose_estimator}
To isolate and evaluate generative diversity, the contact pose estimator maintains strict structural consistency with the force estimator. It shares the identical ViT/DINOv2 backbone to encode the tactile images. The downstream regression MLP head is reconfigured to output a 3-DOF contact pose state: spatial Euclidean coordinates $(x, y)$ and rotation angle $\theta$.

\section{Additional Experiments}\label{app:addition_exp}

\subsection{Physical Consistency and Coverage Validation}
\label{app:physical_validation}

\textbf{Reference-copy calibration.}\quad To test whether the image-quality results can be explained by copying the reference input, we add reference-copy baselines to the evaluation in Table~1. On FeelAnyForce, the baseline obtains an MSE of $233\pm25$ and an SSIM of $0.78\pm0.05$; on AnyTouch2, it obtains an MSE of $81\pm12$ and an SSIM of $0.80\pm0.03$. The substantially larger MSE values than those of \name rule out reference copying as the source of the reported generation quality.

\textbf{Direct force and pose validation.}\quad We further evaluate paired real and generated images using estimators trained only on real data. Across 2,000 paired samples, the force MAE is $0.05\pm0.02$~N for real images and $0.23\pm0.12$~N for generated images. Pose errors are unchanged within measurement variation: across 600 cylinder pairs, the real/generated errors are $8\pm3$/$8\pm2$ pixels and $4\pm2^\circ$; across 200 cross pairs, they are $6\pm2$/$6\pm2$ pixels and $2\pm2^\circ$. These results show that force remains recoverable from generated observations without reducing pose accuracy. The reconstructed-depth results in Figs.~\ref{fig:depth1}--\ref{fig:depth2} and the sensitivity to 1-N force changes in Fig.~\ref{fig:force_var} provide complementary evidence that the generator models geometry and fine force-dependent deformation rather than only RGB brightness.

\begin{figure}[htbp]
    \centering
    \includegraphics[width=0.95\linewidth]{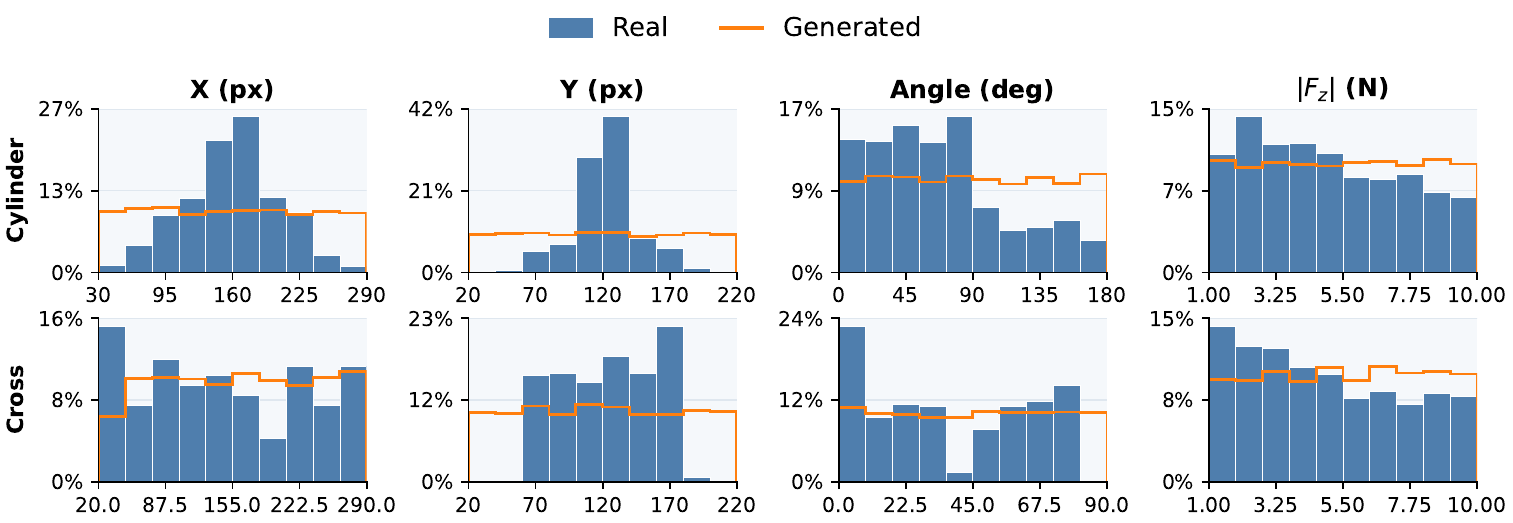}
    \caption{Distribution of pose and force labels in the real and generated training sets. Generated samples fill regions that are sparsely covered by real observations, which is the intended benefit of controllable augmentation.}
    \label{fig:pose_force_coverage}
\end{figure}

\textbf{Coverage-matched comparison.}\quad Dense pose/force coverage is an intended benefit of \name because collecting equivalent real coverage is impractical. Figure~\ref{fig:pose_force_coverage} shows that generated samples fill regions sparsely covered by real training data. To separate this coverage benefit from generation quality, we additionally generate samples at exactly the labels used by the real-data estimator and train a generated-data estimator on this coverage-matched set. The real-data and generated-data estimators obtain identical errors. All real and generated training labels and real test labels use the same manual-alignment procedure, and test poses are unseen in both training sets. Thus, the coverage-matched result rules out estimator degradation caused by generation, while the shared annotation procedure prevents a selective annotation advantage.

\subsection{Manipulation Evaluation Details}
\label{app:manipulation_details}

\textbf{Trial-level uncertainty.}\quad Each imitation-learning setting is evaluated in 25 real-robot trials, a sample count comparable to prior imitation-learning evaluations that use 10--25 trials per setting~\cite{diffusionpolicy, act, dp3}. For the low-data setting, the observed outcomes of 5/25 and 6/25 trials have Wilson 95\% confidence intervals of $[9,39]\%$ and $[12,43]\%$, respectively. Their overlap indicates that this small difference lies within trial-level uncertainty; the more important observation is the consistent improvement at larger demonstration budgets, including $44\%$ versus $28\%$ with 35 demonstrations and $72\%$ versus $48\%$ with 50 demonstrations.

\textbf{Augmentation controls and closed-loop use.}\quad Pose augmentation is not used for the policy because real-time mask labels are unavailable; masks support static pose-conditioned generation only. The policy receives tactile images rather than force values, while force is used solely to condition tactile domain randomization. Figure~\ref{fig:reward} compares 0, 1, 2, 4, and 9 augmented copies, and its non-monotonic trend shows that dataset size alone does not explain the gain. Augmentation is applied at the trajectory level: actions are retained, and a single force offset is shared across every frame of a trajectory. Although individual tactile frames are generated independently, ACT runs at 10~Hz and temporally ensembles overlapping action chunks. Consequently, the closed-loop evaluation directly measures whether force-conditioned tactile augmentation along real trajectories improves policy execution.

\textbf{Failure interpretation.}\quad In low-data regimes, failures arise mainly from insufficient spatial coverage. As the real demonstrations cover more object and target locations, augmentation becomes more effective. This behavior is consistent with the role of \name as an observation-space augmentation method: it introduces contact variation along valid real trajectories but does not create new robot actions or spatial trajectories.

\subsection{Single-Reference Inference and Practical Costs}
\label{app:single_reference}

The term \emph{single reference} describes inference after model pretraining. For each generation request, the inputs are one reference tactile image, a target 3D force, and a target-pose mask. The scalability claim therefore refers to the amount of data that can be generated per reference rather than to training a new generative model from one image. The same architecture supports unseen-object generation and zero-/few-shot transfer to 9DTact. Targets are restricted to the validated force range $F_z\in[1,10]$~N and $F_x,F_y\in[-1,1]$~N; out-of-range targets are excluded.

We also measure the preparation cost of pose conditioning. Constructing an object mask takes less than 3 minutes, and aligning a reference takes less than 10 seconds. One pose mask supports multiple target forces, and an unseen object requires one object mask from which target-pose masks are obtained through rigid transformations. When all real data are already available, performance is near saturation: adding generated samples causes only small changes, and the result varies little for real-to-generated ratios from 1:1 to 1:4 (Table~\ref{tab:mae_results}). In practice, the augmentation ratio can therefore be selected using a real validation set.

\subsection{Comparison with Vision to Tactile Generation}\label{app:vis2tac}
\begin{figure}[htpb]
    \centering
    \includegraphics[width=0.9\textwidth]{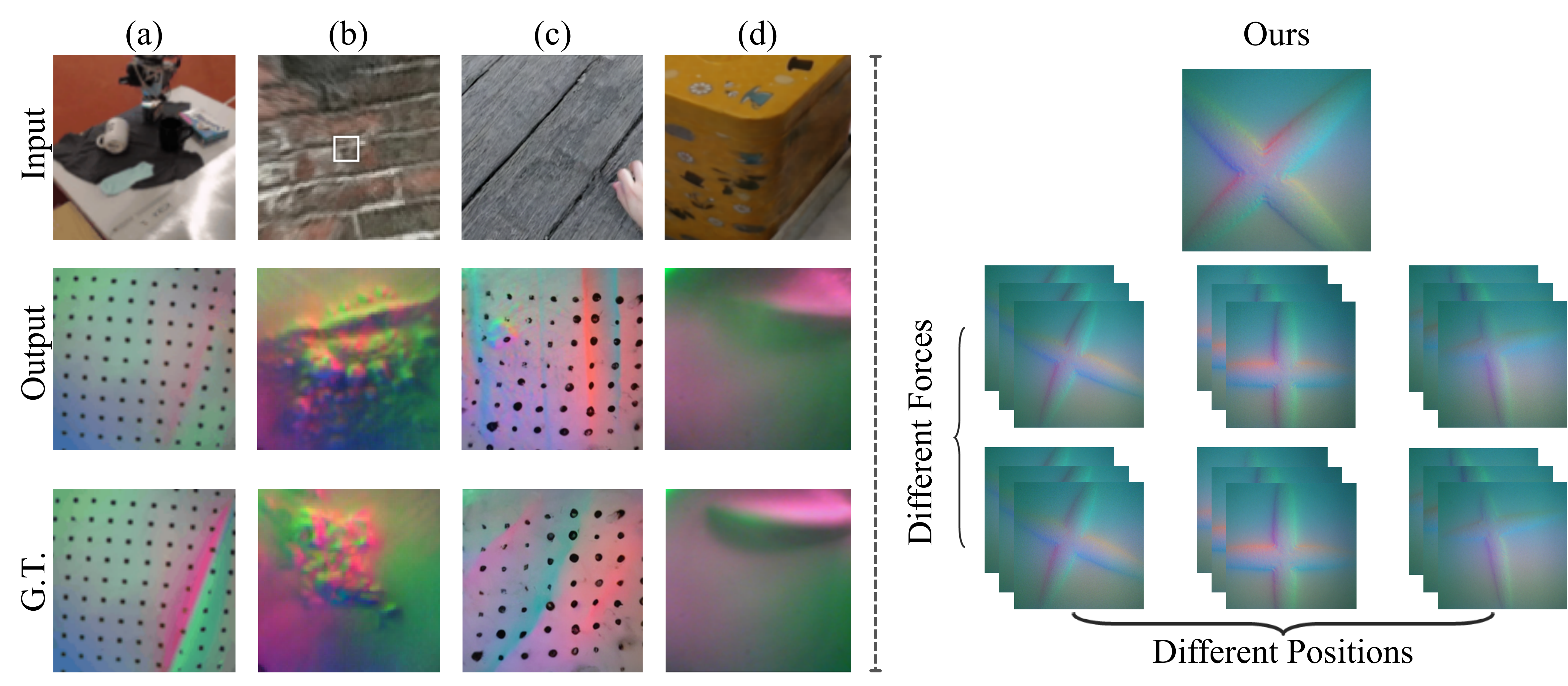}
    \caption{\textbf{Left} refer to different \textsc{Vis2Tac} generation pipeline, where \texttt{(a)} is VisGell~\cite{visgel}, \texttt{(b)} is TaRF~\cite{tarf}, \texttt{(c)} is~\cite{generating}, and \texttt{(d)} is Touching a NeRF~\cite{touchinganerf}. \textbf{Right} refer to our method \name. Compared to other method, with a reference tactile image, \name can generate different images cover various poses and forces.}
\label{fig:vis2tac}
\end{figure}

To highlight the difference between \name and the previous \textsc{Vis2Tac} methods, we demonstrates more examples for different frameworks as shown in Fig.~\ref{fig:vis2tac}. Unlike vision to tactile generation pipeline, which only can generate a single unrealistic image from the reference contact pose, \name can generate various realistic tactile images cover different contraction poses and forces. 

\subsection{Robustness Validation of Contact Mask Alignment}\label{app:mask}

To evaluate the sensitivity of our pose-control module to inaccuracies in contact mask alignment, we conducted a series of controlled experiments by introducing perturbations in three forms: scaling (S), translation (T), and rotation (R). The generation performance was measured using MSE and SSIM metrics, with the quantitative results summarized in Tables~\ref{tab:indiv_perturb} and~\ref{tab:comb_perturb}.

For individual perturbations, scaling the mask within the range of $0.8\times$ to $1.2\times$ produced negligible variations in both MSE and SSIM, suggesting that our method is highly robust to scale changes. Furthermore, translations of up to 4 pixels and rotations of up to $2^\circ$—which correspond to the maximum manual alignment errors typically encountered in practice—resulted in no significant degradation in output quality. Even when pushed to more extreme alignment errors (up to 6 pixels of translation and $6^\circ$ of rotation), the impact on the generated results remained minimal. When scaling, translation, and rotation perturbations were combined simultaneously, the synthesized tactile images remained stable and of high fidelity. 

Crucially, these results highlight the underlying advantage of our decoupled architecture. The contact mask is primarily utilized to specify the spatial contact pose, whereas the actual contact area size and deformation details are governed by the force-control generation (Stage 1). Consequently, manual alignment errors in the mask do not interfere with the force-dependent physical realism, rendering the overall pipeline highly robust to mask noise.

\begin{table}[!ht]
\centering
\renewcommand{\arraystretch}{1.1} 
\caption{Individual Perturbation Analysis with MSE$\downarrow$ and SSIM$\uparrow$, reported as mean ± standard deviation over test samples.}
\label{tab:indiv_perturb}
\resizebox{0.75\textwidth}{!}{
\begin{tabular}{l c >{\centering\arraybackslash}p{2.5cm} >{\centering\arraybackslash}p{2.5cm}}
\toprule
\textbf{Perturbation Type} & \textbf{Parameter} & \textbf{MSE$\downarrow$} & \textbf{SSIM$\uparrow$} \\
\midrule
\multicolumn{4}{c}{\textbf{Scaling (S)}} \\
\midrule
Scaling & 1.0 & 23 ± 2 & 0.83 ± 0.03 \\
Scaling & 1.1 & 23 ± 2 & 0.83 ± 0.03 \\
Scaling & 1.2 & 24 ± 3 & 0.83 ± 0.03 \\
Scaling & 0.9 & 23 ± 2 & 0.83 ± 0.03 \\
Scaling & 0.8 & 24 ± 3 & 0.83 ± 0.03 \\
\midrule
\multicolumn{4}{c}{\textbf{Translation (T)}} \\
\midrule
Translation & 2 px & 23 ± 2 & 0.83 ± 0.03 \\
Translation & 4 px & 24 ± 3 & 0.83 ± 0.03 \\
Translation & 6 px & 25 ± 3 & 0.82 ± 0.03 \\
\midrule
\multicolumn{4}{c}{\textbf{Rotation (R)}} \\
\midrule
Rotation & 2° & 23 ± 2 & 0.83 ± 0.03 \\
Rotation & 4° & 25 ± 3 & 0.82 ± 0.03 \\
Rotation & 6° & 27 ± 3 & 0.82 ± 0.03 \\
\bottomrule
\end{tabular}}
\end{table}

\begin{table}[!ht]
\centering
\small
\renewcommand{\arraystretch}{1.1} 
\caption{Combined Perturbation Analysis with MSE$\downarrow$ and SSIM$\uparrow$, reported as mean ± standard deviation over test samples.}
\label{tab:comb_perturb}
\resizebox{0.7\textwidth}{!}{
\begin{tabular}{l >{\centering\arraybackslash}p{2.5cm} >{\centering\arraybackslash}p{2.5cm}}
\toprule
\textbf{Perturbation Combination} & \textbf{MSE$\downarrow$} & \textbf{SSIM$\uparrow$} \\
\midrule
S 0.9 + T4 + R2 & 23 ± 2 & 0.82 ± 0.03 \\
S 1.1 + T4 + R4 & 26 ± 3 & 0.82 ± 0.03 \\
S 0.8 + T6 + R6 & 28 ± 3 & 0.82 ± 0.03 \\
S 1.1 + T6 + R6 & 28 ± 3 & 0.82 ± 0.03 \\
\bottomrule
\end{tabular}}
\end{table}

\subsection{Details of Object Classification in \textsc{Text2Tac}}\label{app:text2tac_class}
To evaluate \textsc{Text2Tac}~\cite{texttoucher}, we set up a case study to generate 4,800 tactile images with varying contact physical parameters (hardness and texture) for six objects from the TouchToucher training set. 

Images are generated using structured prompt combinations from four groups of tactile descriptors: surface texture (10), stiffness (10), compliance (6), and thermal/frictional properties (8). This yields $10 \times 10 \times 6 \times 8 = 4,800$ unique prompt combinations per object. For instance, for the pill bottle, we employ the template: \textit{``the touch of a pill bottle is \{a\}, \{b\}, \{c\}, \{d\}''}, where placeholders are filled with descriptors from each group (e.g., \textit{``the touch of a pill bottle is smooth, rigid, low compliance, cool''}). 

\begin{figure}[htbp]
    \centering
    \includegraphics[width=1\linewidth]{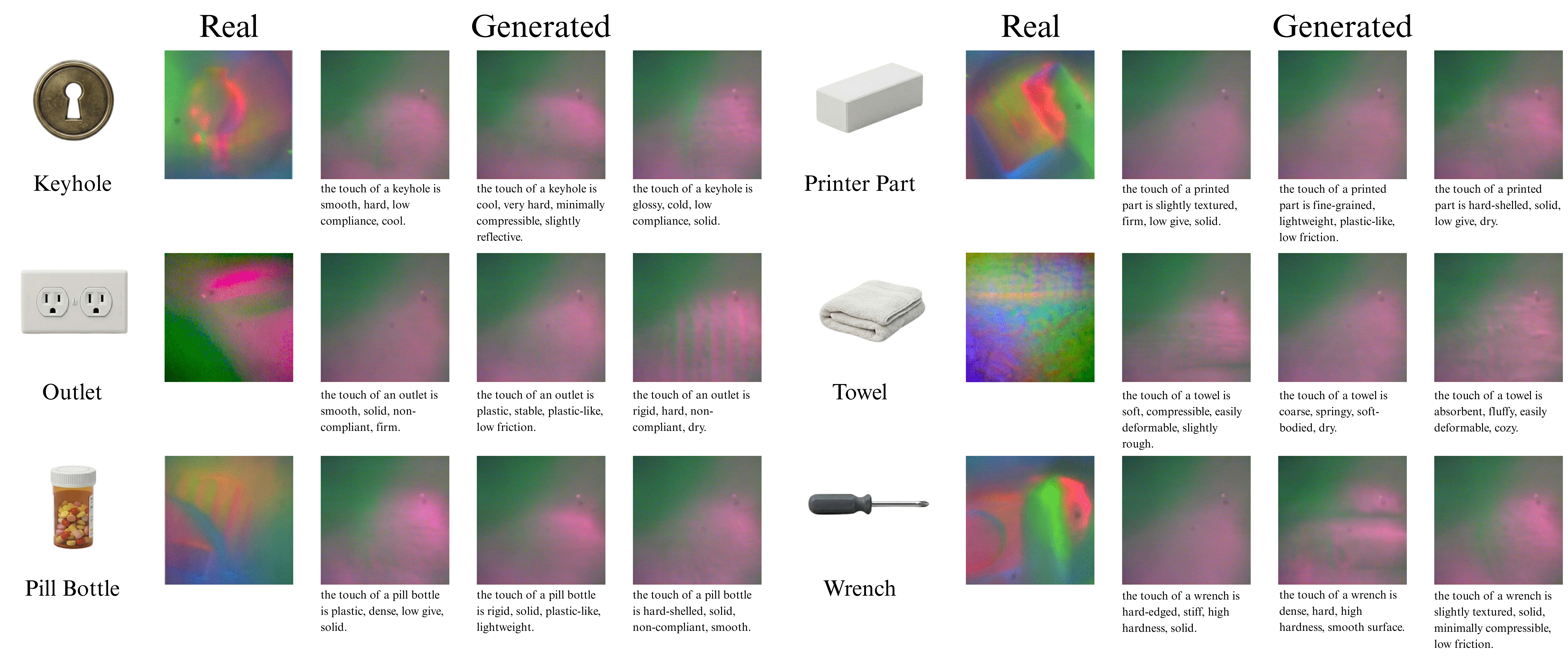}
    \caption{Tactile images of six objects for the classification task. Real: data from the TouchToucher training set; Generated: data produced by \textsc{Text2Tac}~\cite{texttoucher}.}
    \label{fig:text2tac}
\end{figure}

As a reference, a classifier trained on 200 real samples per object achieves an approximate $80\%$ success rate. Visual examples of the real and generated tactile images are shown in Fig.~\ref{fig:text2tac}.

\subsection{Impact of Data Composition on Force Estimation}\label{app:impact}

\begin{table}[!ht]
\centering
\footnotesize
\renewcommand{\arraystretch}{1.2}
\caption{MAE$\downarrow$ of Force Estimator under different data combinations. 
All values are reported as mean ± standard deviation. 
\texttt{Ours} refers to data generated by \name, while \texttt{Real} refers to force estimator trained with real data from FeelAnyForce~\cite{feelanyforce}.}
\label{tab:mae_results}
\resizebox{0.7\textwidth}{!}{%
\begin{tabular}{l >{\centering\arraybackslash}p{3cm} >{\centering\arraybackslash}p{3cm}}
\toprule
\textbf{Training Data Type} & \textbf{Data Size} & \textbf{MAE$\downarrow$} \\
\midrule
\multicolumn{3}{c}{\textbf{Real Data}} \\
\midrule
Real & 10k & 0.061 ± 0.017 \\
Real & 11k & 0.057 ± 0.015 \\
Real & 12k & 0.055 ± 0.015 \\
Real & 13k & 0.051 ± 0.014 \\
Real & 14k & 0.054 ± 0.012 \\
Real & 15k & 0.053 ± 0.012 \\
\midrule
\multicolumn{3}{c}{\textbf{Ours Data}} \\
\midrule
Ours & 15k & 0.21 ± 0.13 \\
Ours & 30k & 0.17 ± 0.09 \\
Ours & 45k & 0.18 ± 0.09 \\
Ours & 60k & 0.16 ± 0.08 \\
\midrule
\multicolumn{3}{c}{\textbf{Real + Ours Data}} \\
\midrule
Real + Ours & 15k + 15k & 0.055 ± 0.021 \\
Real + Ours & 15k + 30k & 0.060 ± 0.017 \\
Real + Ours & 15k + 45k & 0.058 ± 0.015 \\
Real + Ours & 15k + 60k & 0.061 ± 0.014 \\
\bottomrule
\end{tabular}%
}
\end{table}

In our experiments, we observe that training the force estimator with all available real data combined with generated data results in slightly worse performance compared to using real data alone. This phenomenon can be explained from two perspectives.

\texttt{(1)} First, the performance of the force estimator is constrained by the model’s capacity. As shown in Table~\ref{tab:mae_results}, when training with increasing amounts of real data (from 10k to 15k), the MAE improvement plateaus once the dataset exceeds 13k samples, indicating diminishing returns from additional real data.

\texttt{(2)} Second, although the generated data is generally high quality, it inevitably contains minor errors. As the proportion of generated data increases, these errors may accumulate and negatively affect training. Specifically, when training solely on varying amounts of generated data, the MAE fluctuates as the dataset grows, suggesting error accumulation.

Nevertheless, the errors in the generated data are relatively small and do not lead to a substantial degradation in overall performance. This indicates that while excessive generated data may dilute the contribution of real data, it does not fundamentally compromise the results (see Table~\ref{tab:mae_results}). Furthermore, even when substantially larger amounts of generated data (45k or 60k) are combined with real data, the performance does not deteriorate significantly, alleviating concerns regarding the quality of the generated samples.

\subsection{Force Estimation with Force-Control Data Augmentation}\label{app:force}
\begin{figure}[htpb]
\centering
\includegraphics[width=0.75\textwidth]{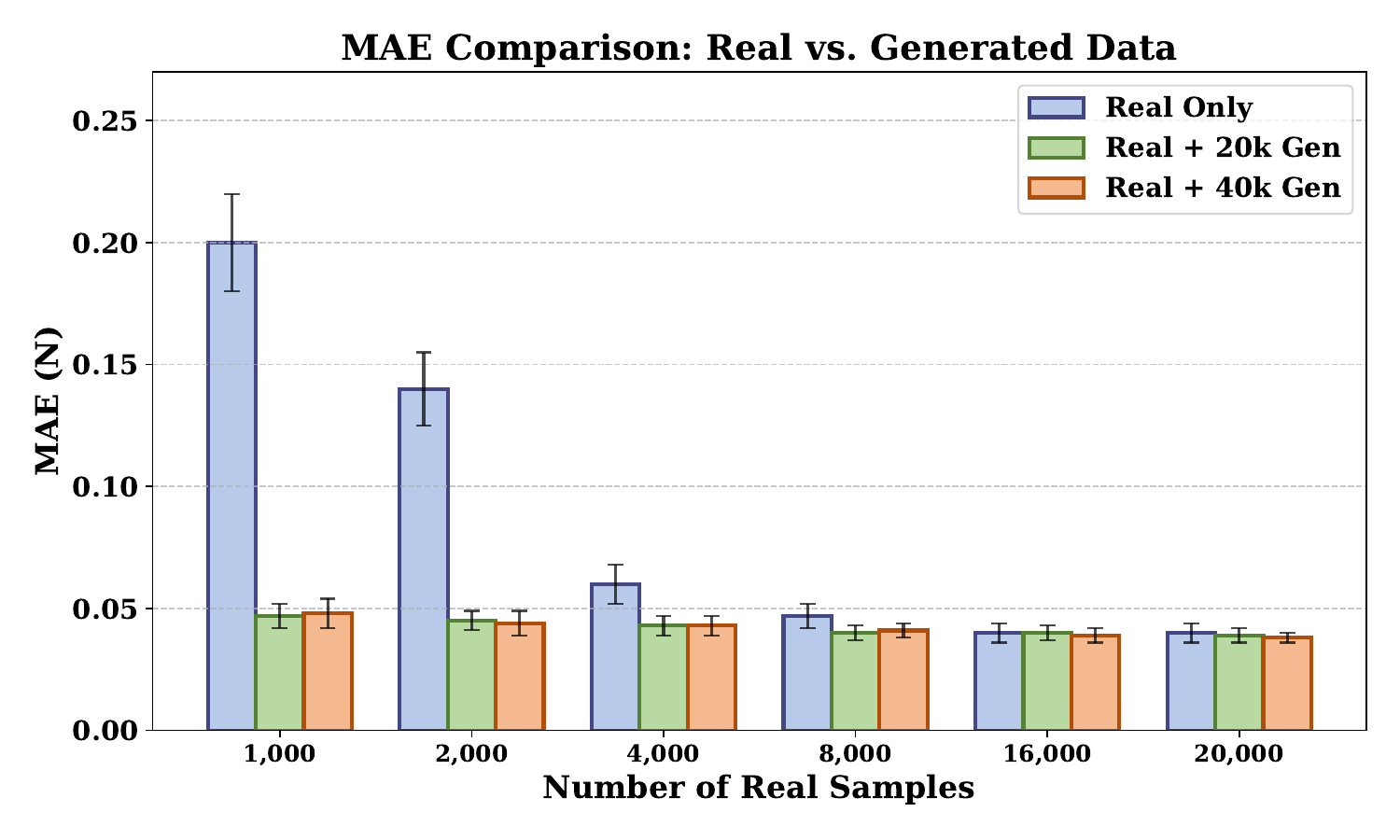}
\caption{Force estimation performance across different training scales. Bars and error bars represent the mean MAE$\downarrow$ and standard deviation (SD) across 1,200 samples, respectively. 
Augmenting small-scale real datasets (1k--4k samples) with 20k or 40k generated samples significantly improves accuracy, effectively bridging the performance gap caused by real-world data scarcity.}
\label{fig:force_est_1}
\end{figure}

To evaluate force-control generation accuracy, we train a force estimator on mixed real and generated data. We augment 1,000 real contact poses with 20 or 40 forces from the generator, yielding 20k and 40k synthetic samples. The estimator is co-trained on 1k–20k real samples combined with these synthetic sets. As shown in Fig.~\ref{fig:force_est_1}, adding generated data reduces MAE when real data is scarce, outperforming real-data-only training. With the generated data, the model matches the full-dataset performance (20k real images) using only 8,000 real samples, indicating that synthetic data enriches force distributions. Combining larger amounts of real and generated data yields the best results, demonstrating the utility of the generated samples.

\subsection{Complete Comparison of Pose Estimation Performance}~\label{app:full_comparison}

\begin{figure}[htpb]
    \centering
    \includegraphics[width=1\linewidth]{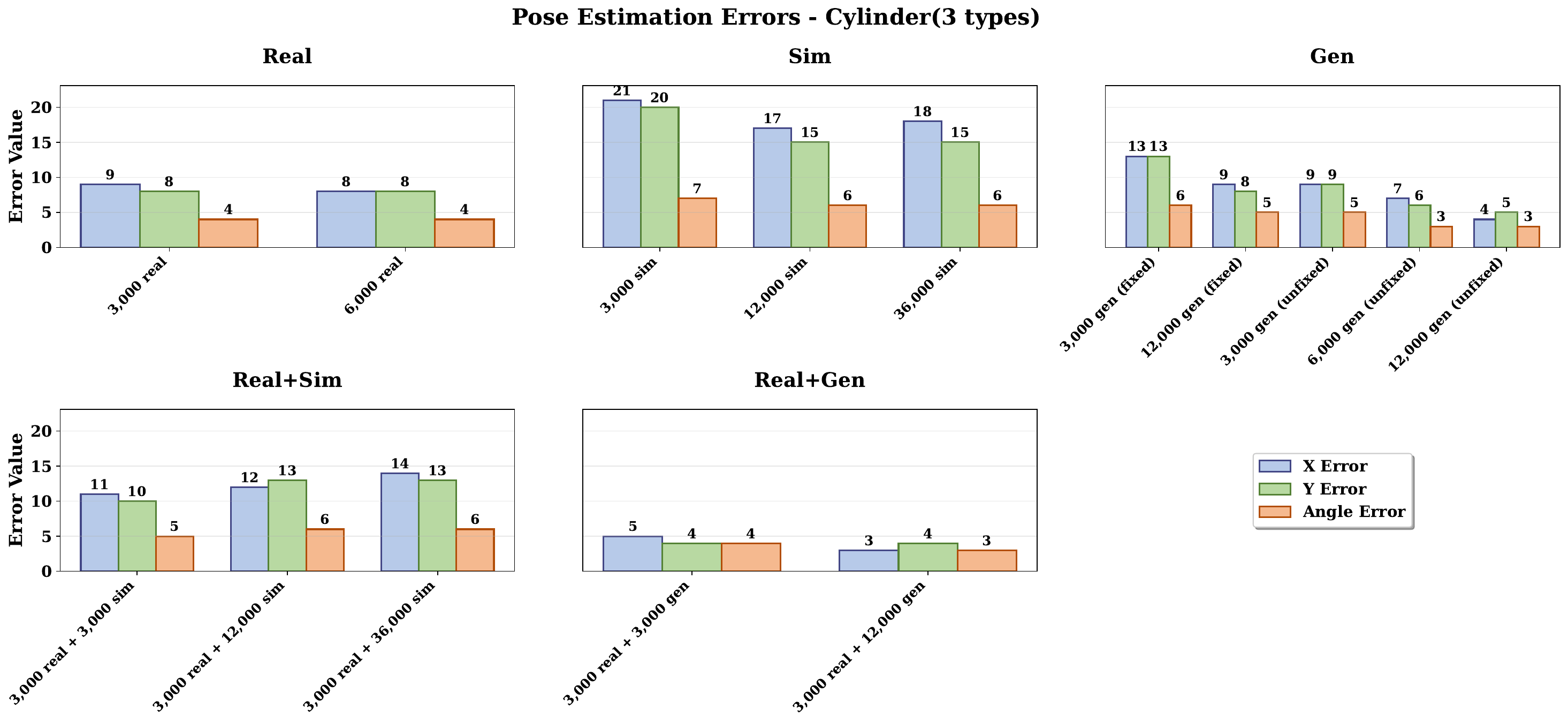}
    \caption{Pose estimation errors (in pixels and degrees) for cylinder objects under different training data regimes, grouped by real, simulated, real+sim, generated, and real+gen datasets. The y-axis is shared across groups to enable scaling comparison. Results are averaged over 5 independent runs; for visual clarity, error bars are omitted as standard deviations for both pixel and angular errors remain consistently within $10\%$ of their respective mean values.}
    \label{fig.cylinder_pose}
\end{figure}

\begin{figure}[htpb]
    \centering
    \includegraphics[width=1\linewidth]{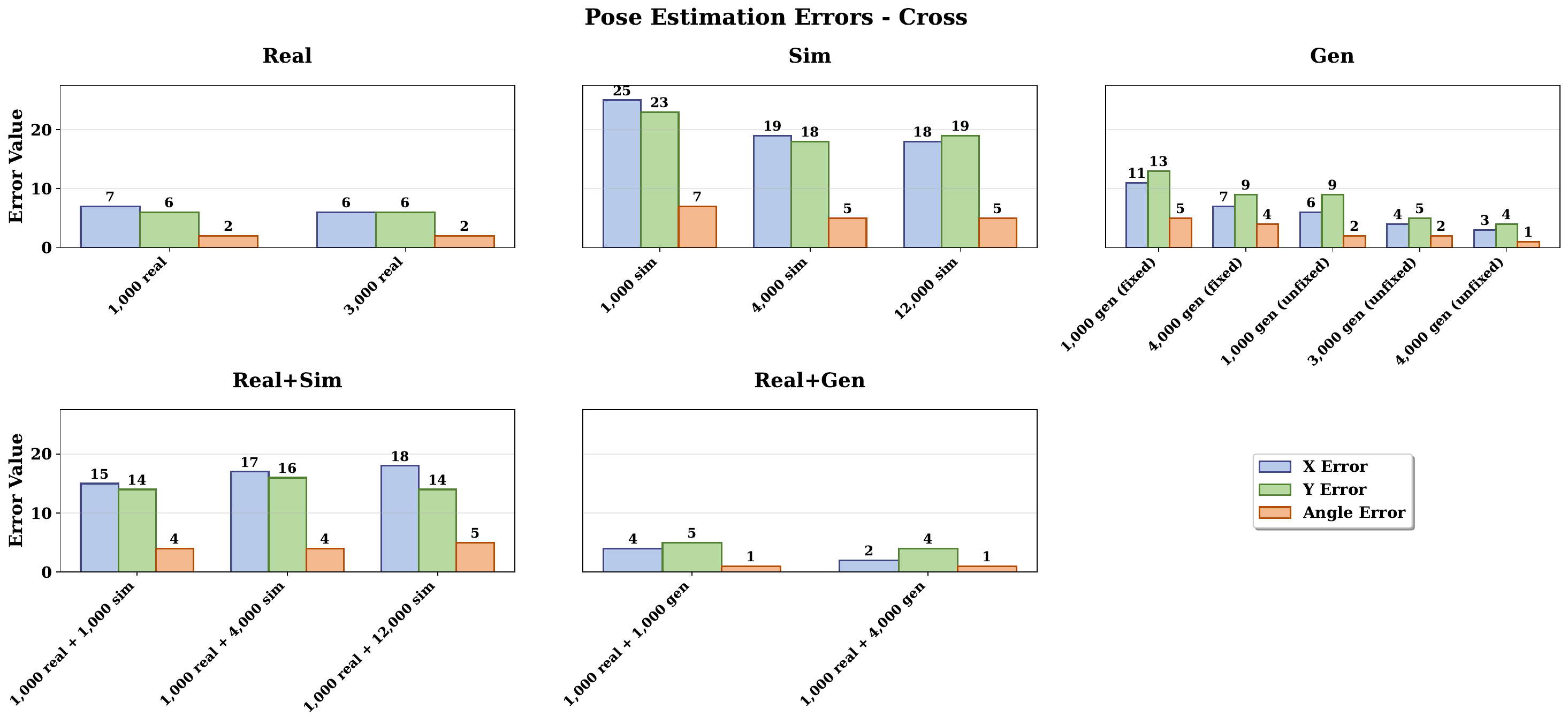}
    \caption{Pose estimation errors (in pixels and degrees) for the cross object under different training data regimes, grouped by real, simulated, real+sim, generated, and real+gen datasets. The y-axis is shared across groups to enable scaling comparison. Results are averaged over 5 independent runs; for visual clarity, error bars are omitted as standard deviations for both pixel and angular errors remain consistently within $10\%$ of their respective mean values.}
    \label{fig.cross_pose}
\end{figure}

\begin{figure}[htpb]
    \centering
    \includegraphics[width=\linewidth]{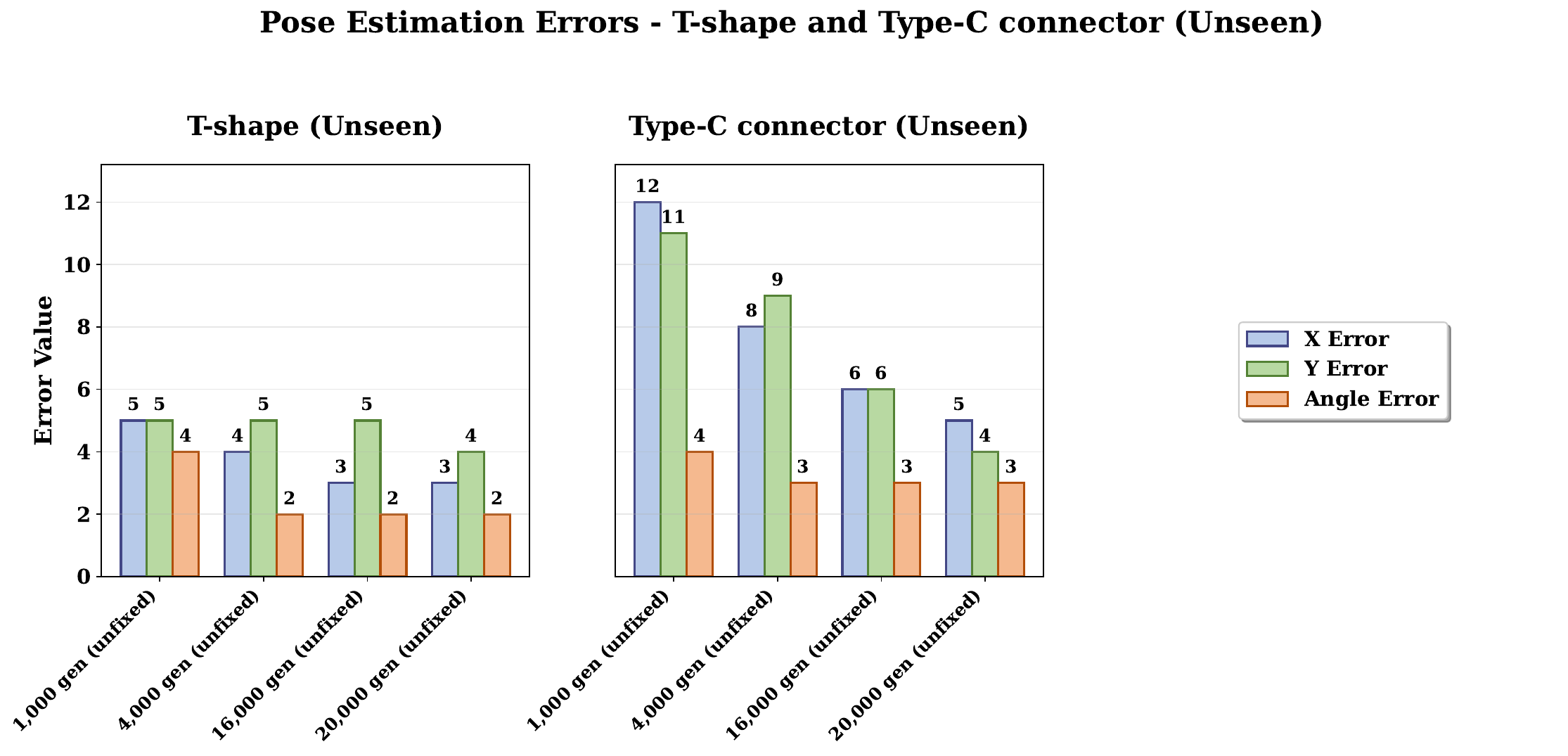}
    \caption{Pose estimation errors (in pixels and degrees) for the unseen T-shape and Type-C connector under different training data regimes, grouped by real, simulated, real+sim, generated, and real+gen datasets. The y-axis is shared across groups to enable scaling comparison. Results are averaged over 5 independent runs; for visual clarity, error bars are omitted as standard deviations for both pixel and angular errors remain consistently within $10\%$ of their respective mean values.}
    \label{fig.tshape_pose}
\end{figure}

\noindent Fig.~\ref{fig.cylinder_pose}, Fig.~\ref{fig.cross_pose} and Fig.~\ref{fig.tshape_pose} provide a comprehensive comparison of pose estimation errors under different training data settings, measured in pixels for X and Y coordinates and in degrees for orientation. The results are reported for four object categories: Cylinder (with three types), Cross, unseen T-shape, and the Type-C connector. Overall, pose estimators trained solely on generated data achieve strong performance and often outperform models trained on real data, even when the real dataset is relatively large. Simulation-based data from Taxim~\cite{taxim} consistently underperforms compared to both real and generated data, and combining real with simulation data generally offers limited improvements or even degrades performance due to the lower realism of simulated samples. Using varying force during data generation further improves performance over fixed-force generation since it can cover more variations in the real-world scenario. Notably, for the unseen T-shape object and Type-C connector, models trained on generated data generalize effectively, demonstrating the value of generated tactile images in covering diverse contact poses and angles.

\subsection{Details of Object Weighting}\label{app:push}

\noindent Table~\ref{tab:object_pushing} summarizes the evaluation of the force estimator on the object-pushing task. The objective of this downstream task is to evaluate whether a force estimator trained on ControlTac-generated data can achieve performance comparable to one trained on real data, thereby highlighting the high quality of the data generated by ControlTac. The table provides two sets of numbers for each object; for instance, in the entry “Bottle (0.63)”, 0.63 refers to the mass of the glass bottle in kilograms. The other numbers indicate the measured and predicted pushing forces: $1.08\,\text{N}$ is the force measured by the force sensor, $1.14\,\text{N}$ is the force predicted by the estimator trained on real-world data, and $1.16\,\text{N}$ is the force predicted by the estimator trained on ControlTac-generated tactile data. During the pushing process, the applied force is dynamic: as the robotic arm pushes the object at a constant speed, force sensor readings and corresponding tactile data are recorded and used by the estimator to predict the pushing force. 

\begin{table}[!htpb]
\centering
\renewcommand{\arraystretch}{1.2} 
\caption{Results of object pushing experiments for the four objects. 
All values represent the measured force in Newtons (N), reported as mean ± standard deviation.}
\label{tab:object_pushing}
\setlength{\tabcolsep}{6pt} 
\resizebox{0.9\linewidth}{!}{%
\begin{tabular}{l >{\centering\arraybackslash}p{2.5cm} >{\centering\arraybackslash}p{2.5cm} >{\centering\arraybackslash}p{2.5cm} >{\centering\arraybackslash}p{2.5cm}}
\toprule
Force [N] & Weight (1.0) & Cyl. (0.50) & Cyl. (0.56) & Bottle (0.63) \\
\midrule
Force ATI (G.T.) & 2.24 & 0.96 & 1.06 & 1.08 \\
Force (Real Data) & 2.38 ± 0.35 & 1.08 ± 0.14 & 1.18 ± 0.17 & 1.14 ± 0.23 \\
Force (Ours) & 2.36 ± 0.45 & 1.11 ± 0.21 & 1.17 ± 0.19 & 1.16 ± 0.27 \\
\bottomrule
\end{tabular}%
}
\end{table}

Across all four objects—including the  calibration weight, beaker filled with water, and glass bottle—the estimator trained on generated images achieves performance comparable to that of the real-data-trained model. This demonstrates that training with generated data enables the force estimator to generalize effectively to diverse objects with varying textures, materials, and weights, closely matching the accuracy of the real-data-trained model.

\subsection{Details of Object Insertion}\label{app:insertion}
For the robotic insertion tasks---comprising a precise insertion task using a hole (5~cm in height, 3~cm in depth, and 10~mm in diameter) alongside three 3D-printed objects (a 7-cm-long cylinder, a $7\times3$~cm cross-shaped object, and a $7\times3$~cm T-shaped object, all with a 7~mm diameter) and a USB insertion task involving a Type-C cable and a charger---an XArm7 robot equipped with two GelSight Mini tactile sensors is deployed to grasp the target object at a random angle and in-hand pose above a known hole location. To successfully complete the insertion, the robot dynamically adjusts its end-effector pose based on tactile feedback: the pose estimator first predicts the object's pose on the tactile sensor, and the Euclidean distance from the estimated pose to the center is computed and converted from pixel units to real-world dimensions using a scaling factor of 1~pixel = $\frac{1}{20}$~mm. The robot then compensates for the initial grasping uncertainty by rotating along the $R_x$ axis and translating along the $y$ axis, thereby vertically aligning the object directly above the hole for insertion. As a practical real-robot application, this task evaluates errors in both position and orientation; furthermore, since insufficient data quality degrades estimator performance and compromises operational reliability, continuously monitoring the performance of the pose estimator serves as an effective strategy to mitigate potential safety hazards during execution.

\subsection{Details of Imitation Learning}\label{app:imitation}
\begin{figure}[htpb]
    \centering
    \includegraphics[width=0.6\linewidth]{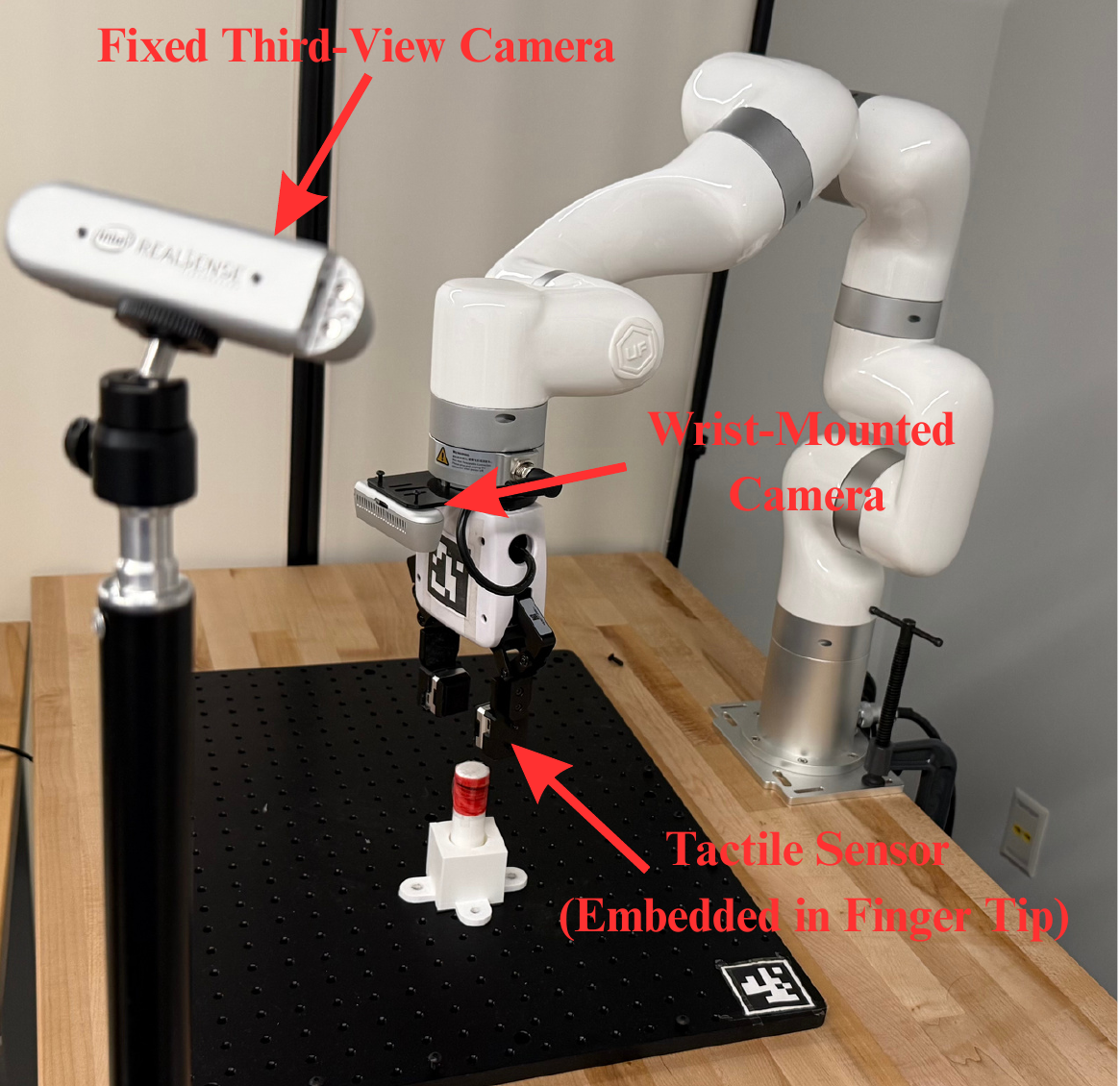}
    \caption{The real-world experimental setup for the Pick and Peg-in-Hole task. The hardware system consists of a 7-DoF xArm7 robot arm, a parallel gripper equipped with an integrated tactile sensor, a third-view camera for global workspace monitoring, and a compact wrist-mounted camera for localized hand-eye coordination.}
    \label{fig:robot_setup}
\end{figure}

For the real-world imitation learning task (Pick and Peg-in-Hole task), our experimental platform is structured as shown in Fig.~\ref{fig:robot_setup}. We deploy our visual-tactile policies on a 7-DoF xArm7 robot arm equipped with a parallel gripper, operated at a control frequency of 10~Hz. The multi-modal observation space integrates sensory data from three distinct sources: a fixed third-view RGB camera overlooking the workspace, a wrist-mounted RGB camera capturing localized visual feedback, and a tactile sensor embedded within the gripper fingers to capture contact interaction profiles.

\begin{figure}[htpb]
    \centering
    \includegraphics[width=0.55\linewidth]{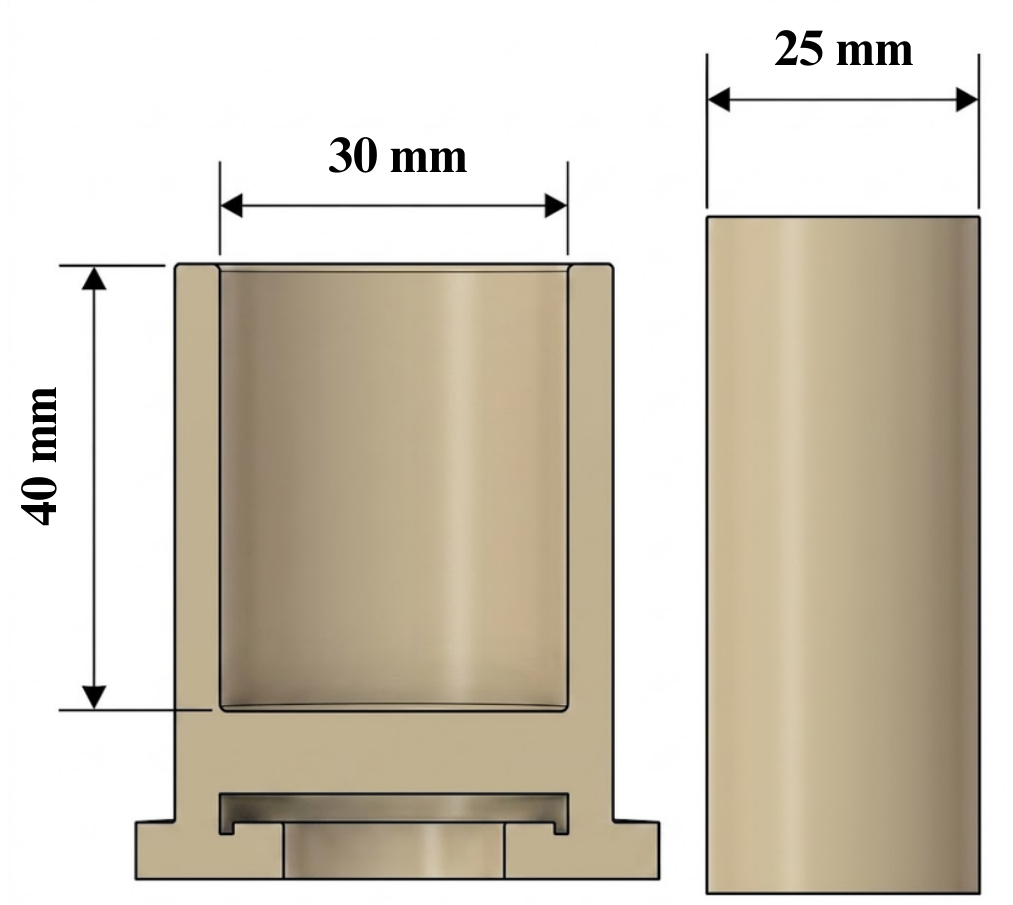} 
    \caption{Detailed geometric dimensions of the peg and hole components. The cylindrical hole (left) has an inner diameter of 30 mm and an insertion depth of 40 mm, while the cylindrical peg (right) has an outer diameter of 25 mm.}
    \label{fig:dimensions}
\end{figure}

To rigorously evaluate the insertion performance under precise tolerance constraints, we utilize a standardized peg-and-hole assembly with specific geometric dimensions, as illustrated in Fig.~\ref{fig:dimensions}. The cylindrical peg features an outer diameter of $25\text{ mm}$. The corresponding mating component consists of a cylindrical receptacle (hole) with an inner diameter of $30\text{ mm}$ and an effective insertion depth of $40\text{ mm}$. This clearance configuration requires the policy to accurately align the mating axes before applying downward force, making it highly sensitive to contact feedback.

We train the Action Chunking Transformer (ACT)~\cite{aloha} on the collected multi-modal dataset. Specifically, ACT encodes the visual observations, tactile inputs, and robot proprioception using a transformer-based conditional VAE, and subsequently predicts a chunk of $N_a = 20$ future actions instead of a single action. We optimize the network using the standard ACT training objective, which comprises the reconstruction loss (L1 loss) for actions and the KL regularization term for the latent variable. During test time, we implement asynchronous inference to ensure real-time deployment, and apply temporal ensembling over overlapping action chunks to reduce action jitter. Although ACT can effectively learn from relatively small demonstration datasets (20--100 real demonstrations in our setup), its action predictions can become noisy for contact-rich tasks when the teleoperation data exhibits inconsistent contact behaviors. This vulnerability makes ACT an excellent stress test for our force-controlled tactile data augmentation pipeline, allowing us to demonstrate how tactile augmentation significantly enhances policy robustness against such contact inconsistencies.

\section{Failure Analysis}
\label{app:failure}

\begin{figure}[htpb]
    \centering
    \includegraphics[width=0.9\linewidth]{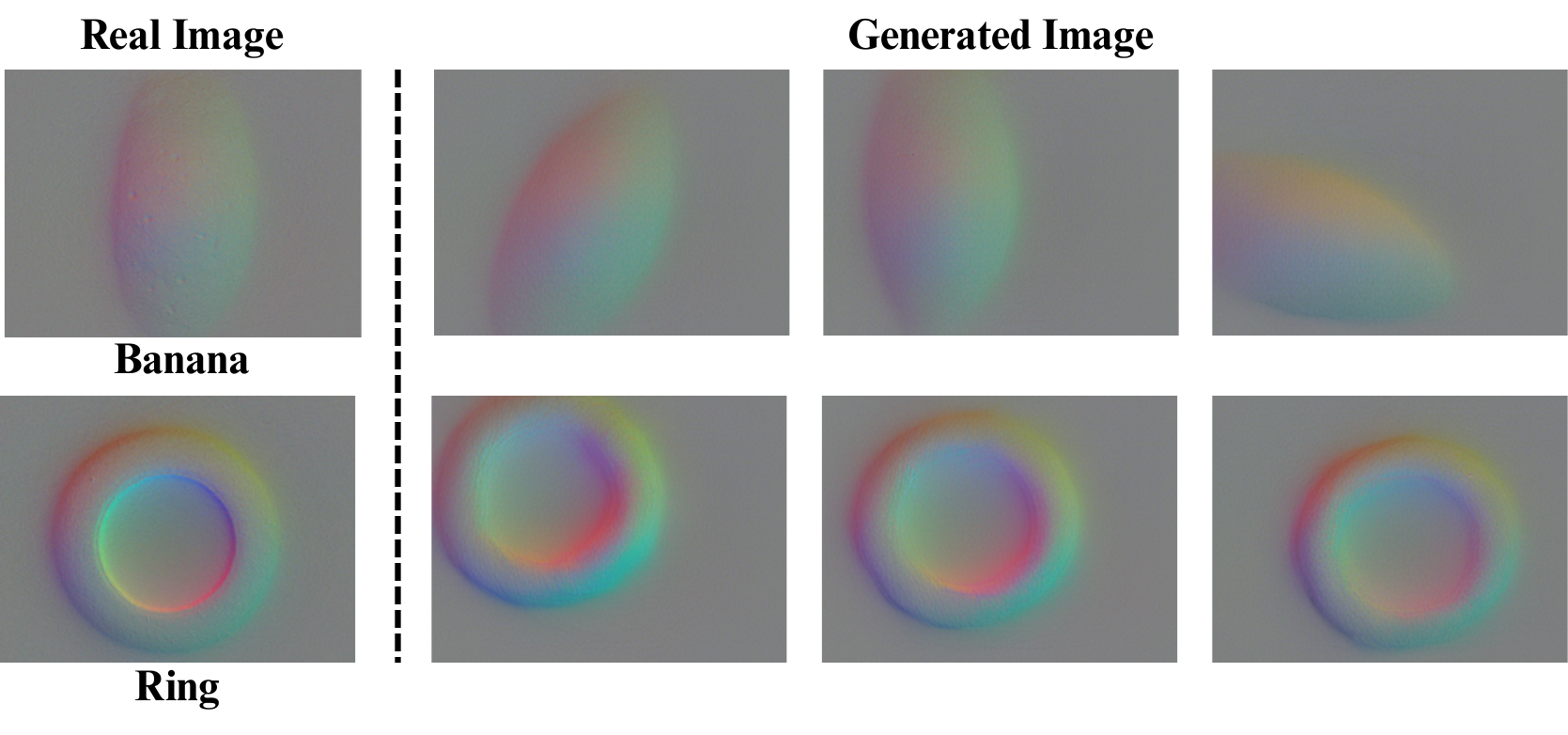}
    \caption{Visualization of failure cases (banana and flattened ring). Background is subtracted and a constant offset of 127 is applied to normalize pixel values for display.}
    \label{fig.failure}
\end{figure}

The model's generalization is primarily constrained by the homogeneity of the training set, which exclusively features 3D-printed PLA objects with a narrow range of curved surfaces. Consequently, performance bottlenecks emerge when encountering complex textures, varying material hardness, or surface curvatures that deviate from the training distribution, such as the flattened ring in Fig.~\ref{fig.failure}. To address these limitations in geometric and material diversity, we augmented the 20,000-sample dataset with 1,000 targeted tactile images of an object with different curvature (e.g., a flat-surfaced cube). This strategic addition significantly improved reconstruction for previously unseen geometries—reducing the MSE from 35 to 27 and increasing the SSIM from 0.80 to 0.83—demonstrating that even a modest infusion of domain-specific data can effectively bridge performance gaps in underrepresented tactile scenarios.

\section{Addition Visualizations}\label{app:vis}
In this section, we provide additional visualizations to clarify the concepts discussed.

\subsection{Complete Visualization Results} \label{app:full_vis}

\begin{figure}[htpb]
    \centering
    \includegraphics[width=\textwidth]{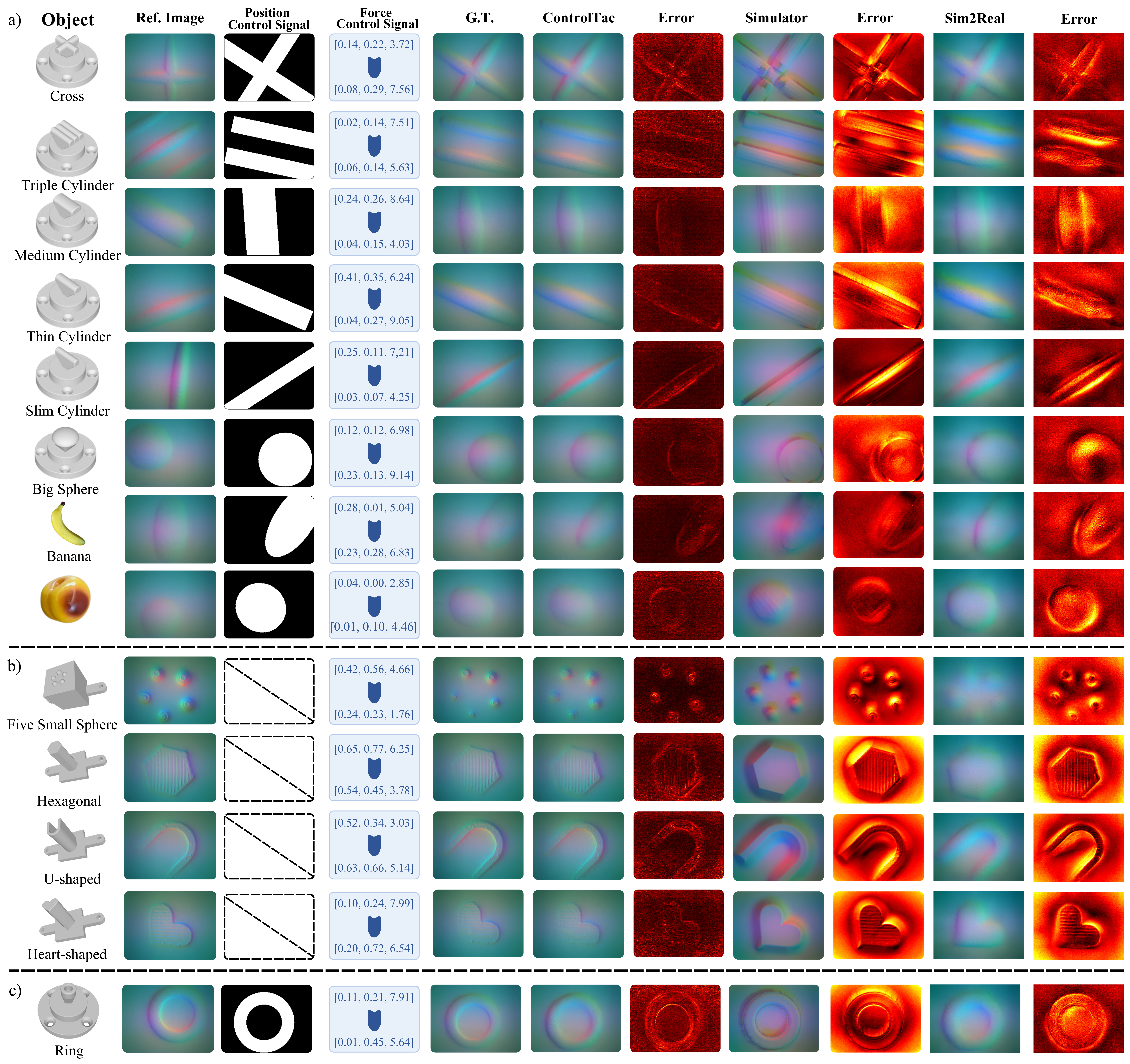}
    \caption{\textbf{Comprehensive qualitative generation results across diverse objects and contact conditions.}
    Each row displays the object's 3D preview, input tactile image, contact mask, and initial/target forces, followed by the Ground Truth, our \name, \textsc{Simulator}~\cite{taxim, difftactile, taccel}, and \textsc{Sim2Real}~\cite{sim2real} outputs with their respective error maps. 
    \textbf{(a)} Seen objects (rows 1--6) and unseen objects (rows 7--8) under novel poses/forces from FeelAnyForce~\cite{feelanyforce}. 
    \textbf{(b)} Unseen objects from AnyTouch2~\cite{anytouch2}. 
    \textbf{(c)} A failure case on an unseen object from FeelAnyForce~\cite{feelanyforce}.}
    \label{fig:vis_all_full}
\end{figure}

In this section, Fig.~\ref{fig:vis_all_full} presents the comprehensive qualitative generation results of \name across diverse objects and contact conditions under novel poses and forces. The figure displays the detailed 3D previews, inputs, and generated tactile outputs along with their corresponding error maps for both seen and unseen physical scenarios.

\begin{figure}[htpb]
    \centering
    \includegraphics[width=1\textwidth]{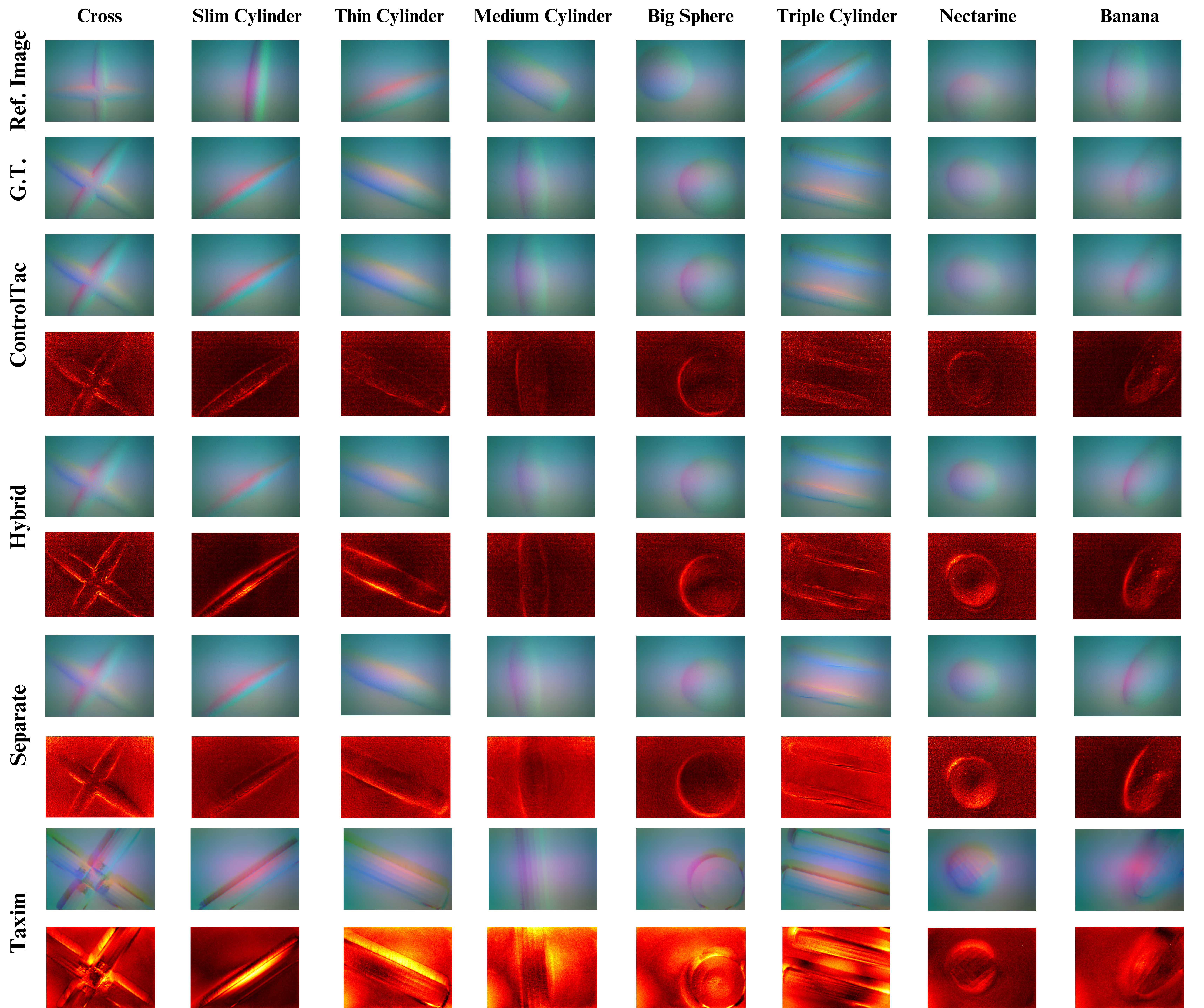}
    \caption{Full visualization results comparing \name with two baseline models and a simulator~\cite{taxim, difftactile, taccel}.}
    \label{fig:error_map}
\end{figure}

Furthermore, Fig.~\ref{fig:error_map} presents the full visualization results of \name compared with two baseline models and a simulator~\cite{taxim, difftactile, taccel}. The figure illustrates performance across multiple scenarios, showing that \name consistently achieves more accurate and robust predictions than the baselines

\subsection{Visualization of Generated Image using Force-Control Generation}\label{app:force_only}

\begin{figure}[htbp]
    \centering
    \includegraphics[width=0.8\textwidth]{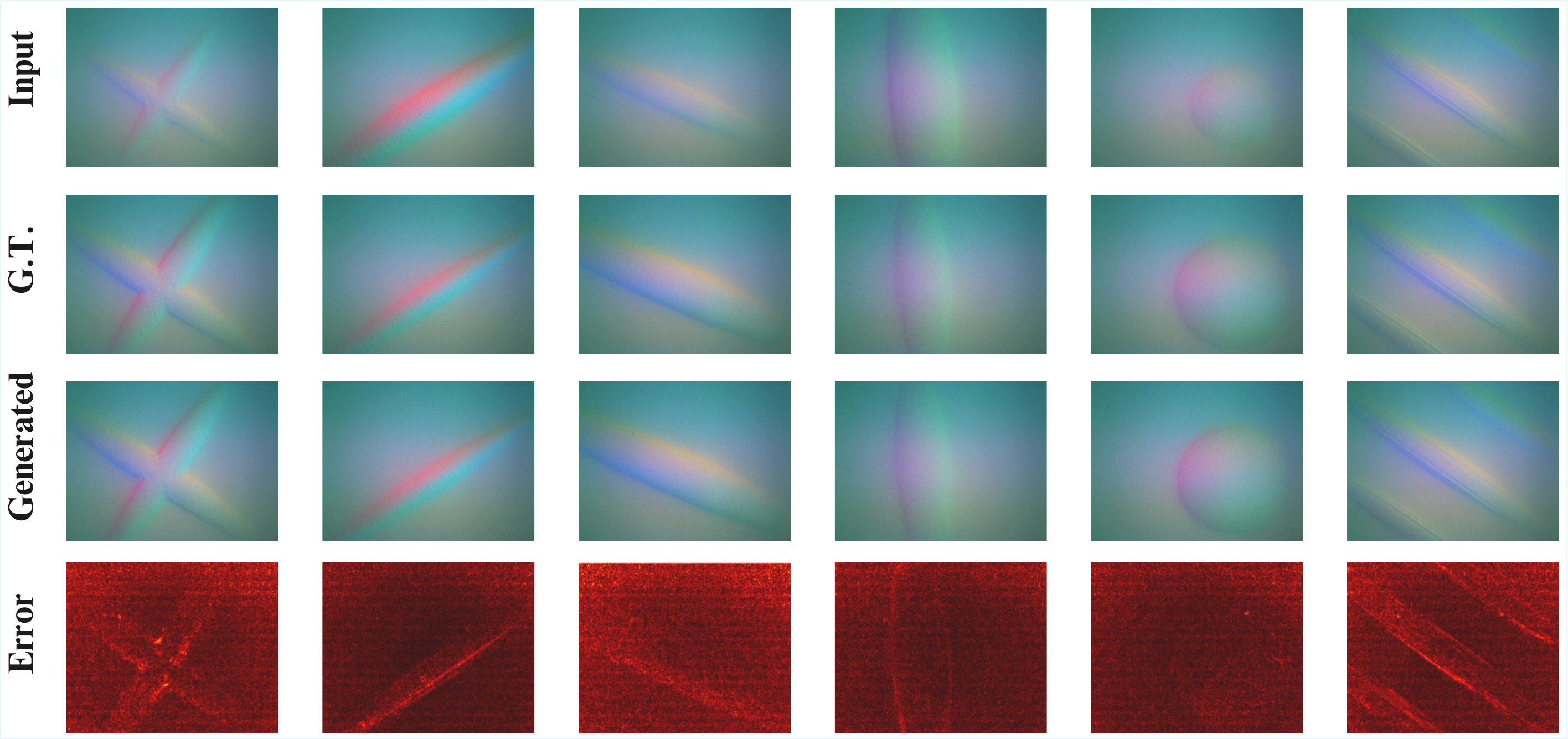}
    \caption{Generated tactile images using the force-control generation component of \name at the same contact pose.}
    \label{fig:only_force_gen}
\end{figure}

In this section, we showcase the visualization results using the force-control generation component of \name. Fig.~\ref{fig:only_force_gen} presents the generated tactile images for the same contact pose, demonstrating excellent results and the effectiveness of this component.

\subsection{Visualization of Depth Map}\label{app:depth}

\begin{figure}[htpb]
    \centering
    \includegraphics[width=0.8\textwidth]{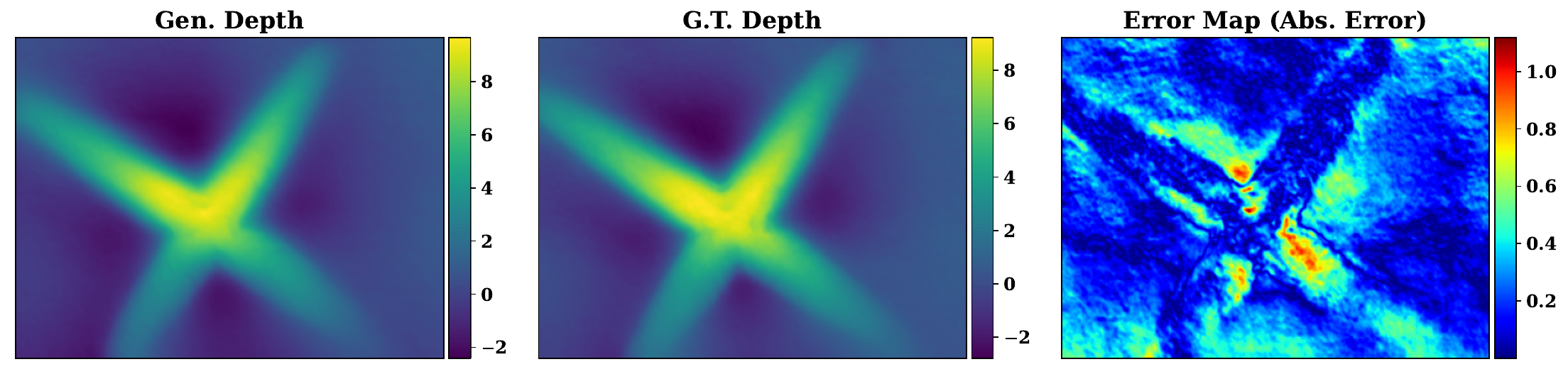}
    \caption{Depth maps of the Cross object generated from \name-generated tactile images (Gen.), compared with Ground Truth (G.T.) and corresponding error maps.}
    \label{fig:depth1}
\end{figure}

\begin{figure}[htpb]
    \centering
    \includegraphics[width=0.8\textwidth]{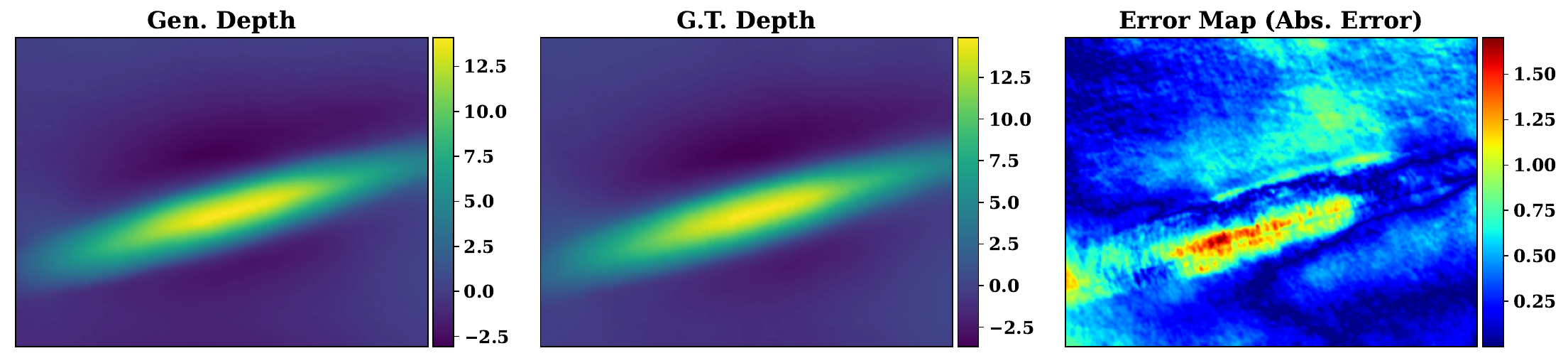}
    \caption{Depth maps of the Thin Cylinder object generated from \name-generated tactile images (Gen.), compared with Ground Truth (G.T.) and corresponding error maps.}
    \label{fig:depth2}
\end{figure}

Since standard RGB tactile images cannot fully capture the physical deformation of the gel and the corresponding force distribution, we further evaluate the depth derived from the tactile images generated by \name. Figures~\ref{fig:depth1} and~\ref{fig:depth2} present these depth visualizations alongside the corresponding ground truth and depth error maps. These results confirm that \name accurately captures the underlying surface geometry and physical deformation, verifying that our generator does not merely scale the overall image brightness.

\subsection{Object and Tactile Image Visualization for Classification}\label{app:class}
\begin{figure}[htbp]
    \centering
    \includegraphics[width=1\linewidth]{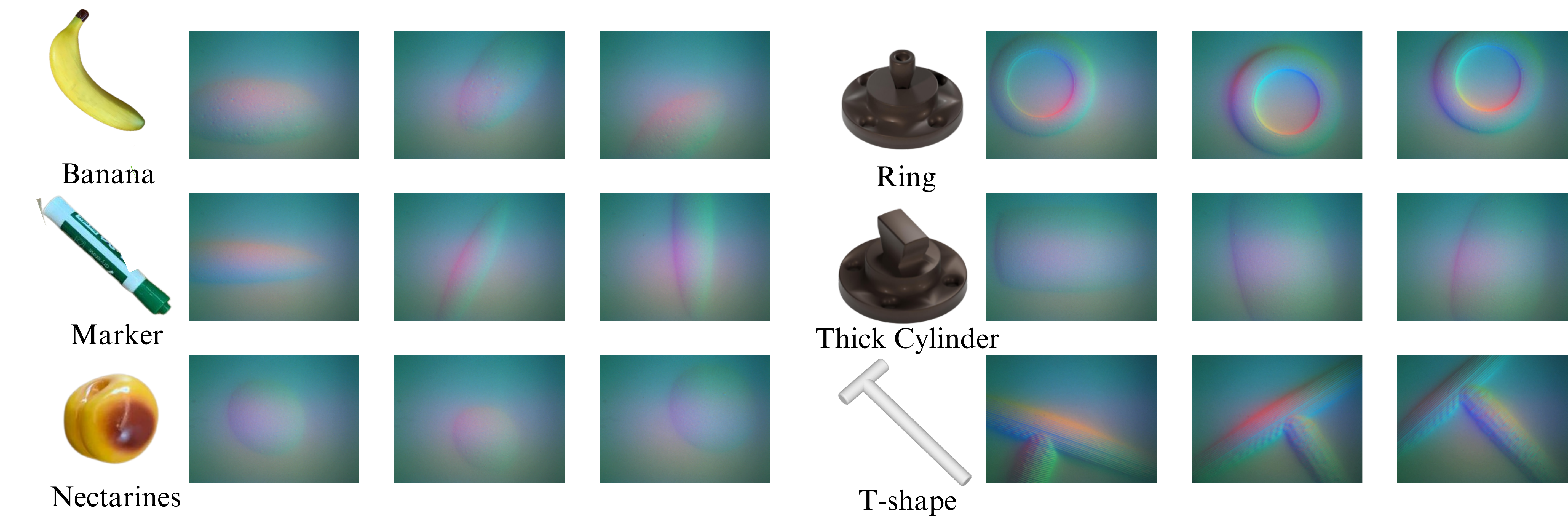}
    \caption{Six objects and their corresponding tactile images used in the classification task.}
    \label{fig:class}
\end{figure}

In this section, we present six objects used in the classification task along with their corresponding tactile images, as shown in Fig.~\ref{fig:class}.

\subsection{Sensitivity of Tactile Images to Force Variations}
\label{app:force_sensitivity}

\begin{figure}[h]
    \centering
    \includegraphics[width=0.8\textwidth]{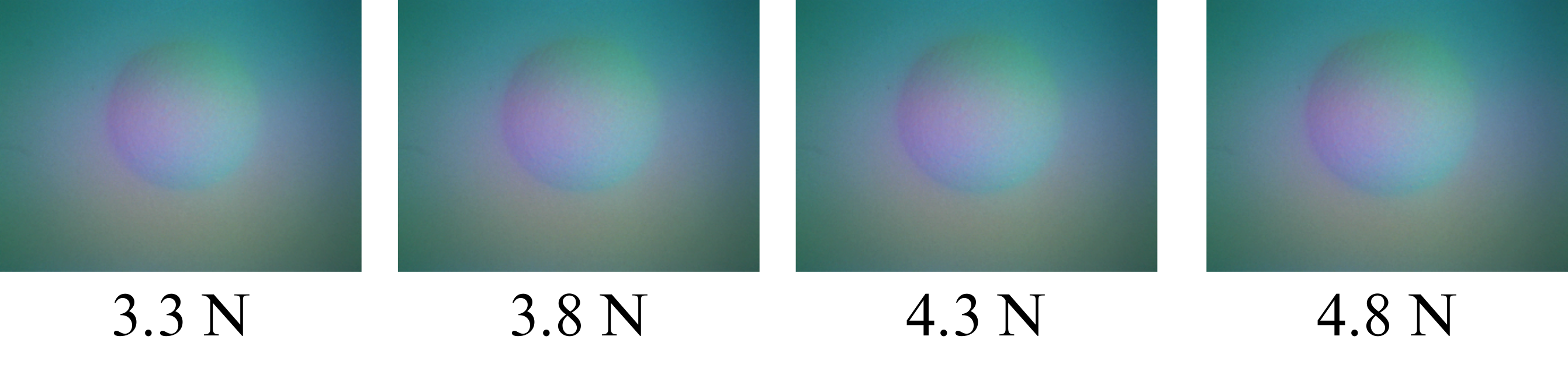} 
    \caption{Tactile images corresponding to four different force values: 3.3\,N, 3.8\,N, 4.3\,N, and 4.8\,N.}
    \label{fig:force_var}
\end{figure}

Fig.~\ref{fig:force_var} illustrates the high sensitivity of the force estimation task to tactile image accuracy. We selected four force values—3.3\,N, 3.8\,N, 4.3\,N, and 4.8\,N—to show how subtle force differences affect tactile images. Differences smaller than 0.5\,N are almost imperceptible, and even differences of 1\,N result in only very subtle changes. This highlights the difficulty of accurately labeling tactile images with corresponding forces and underscores the challenge of using generated or simulated tactile images for this task.

\subsection{Visualization of different sensor samples}\label{app:sensor}
\begin{figure}[htpb]
    \centering
    \includegraphics[width=0.60\textwidth]{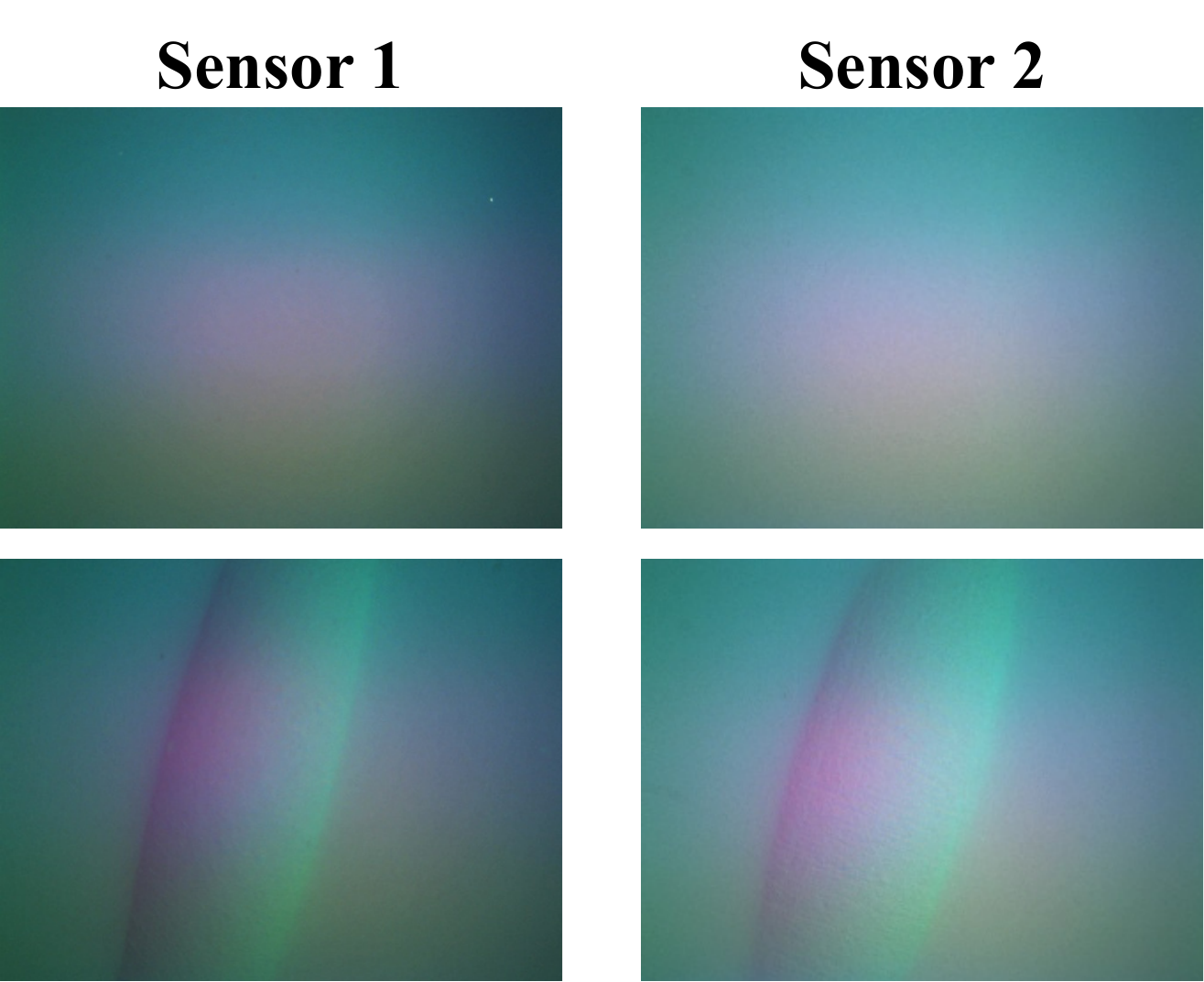} 
    \caption{The background of Sensor 1 is noticeably darker than that of Sensor 2.}
    \label{fig:sensor}
\end{figure}

In this section, we visualize the difference of contact image between different sensor samples, as shown in Fig.~\ref{fig:sensor}.

\subsection{Visualization of a special case requiring generation}\label{app:special}
\begin{figure}[h]
    \centering
    \includegraphics[width=0.8\textwidth]{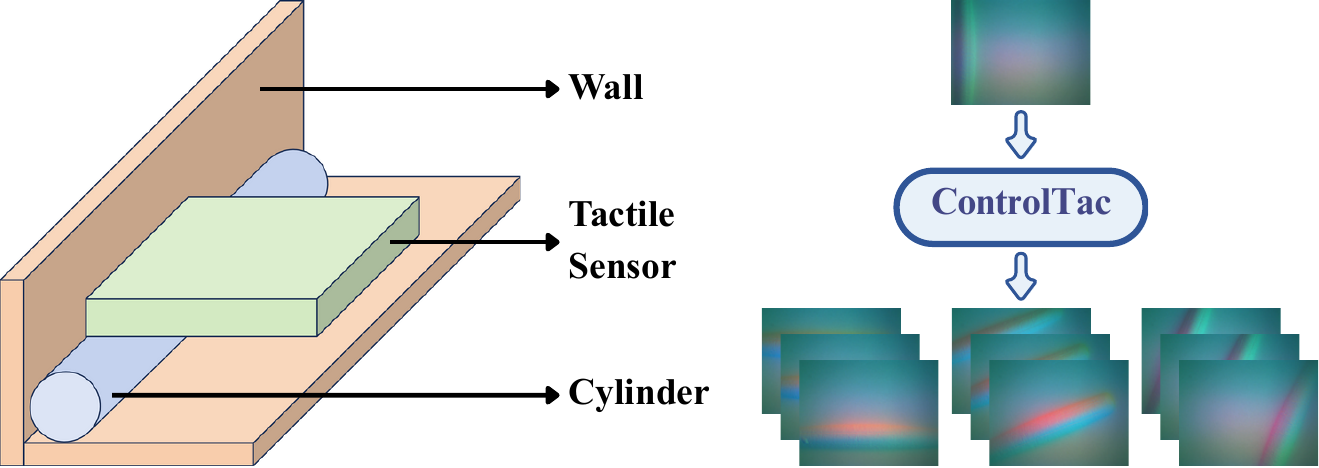} 
    \caption{With the cylinder placed in the corner, the sensor only can get several images contact with the edge. In this case, \name can generate various images with more contact poses.}
    \label{fig:wall}
\end{figure}

In this section, Fig.~\ref{fig:wall} visualize a simple special case, where the cylinder is placed in the corner. In this case, the sensor is only able to contact with the cylinder with the edge of the sensor, where it cannot cover the other contact poses. Notably, it's simple conditions, whereas it easily to happen in the real world with some articulate objects like a cabinet or a action figure.



\subsection{Visualization Results in the Simulator}
\label{app:vis_taxim}

\begin{figure}[htbp]
    \centering
    \includegraphics[width=0.95\linewidth]{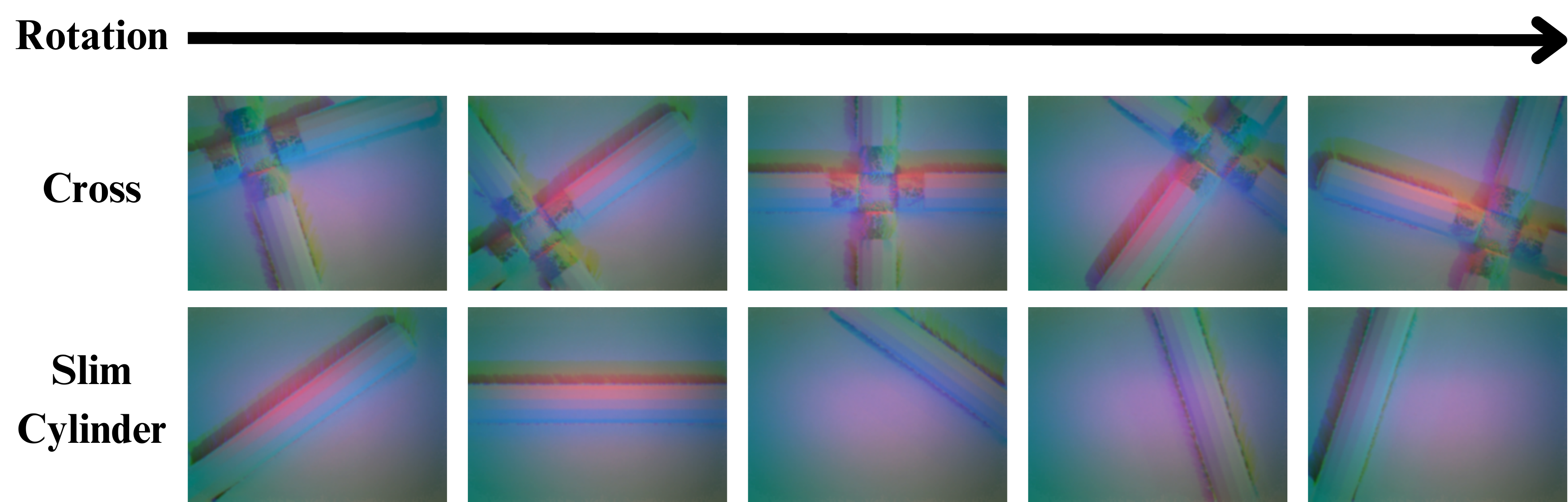} 
    \caption{Simulated tactile images using Taxim~\cite{taxim}.}
    \label{fig:taxim}
\end{figure}

In this section, we present tactile images generated using a simulator~\cite{taxim,taccel,difftactile}. As shown in Fig.~\ref{fig:taxim}, the simulated images lack realism, highlighting the limitations of current simulation methods for tactile data.

\subsection{Visualization of the Type-C Connector Insertion Task}
\label{app:usb}

\begin{figure}[htpb]
    \centering
    \includegraphics[width=0.7\linewidth]{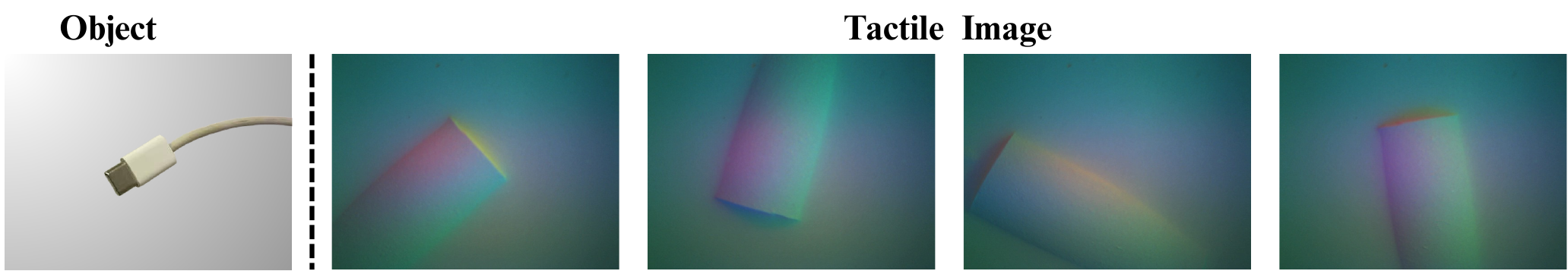}
    \caption{Real USB Type-C connector and corresponding tactile image.}
    \label{fig.usb}
\end{figure}

To better illustrate the Type-C connector insertion task, Fig.~\ref{fig.usb} shows both the real Type-C connector and its corresponding tactile image used in this experiment.

\clearpage

\end{document}